\documentclass[letterpaper, 10 pt, conference]{ieeeconf}
\IEEEoverridecommandlockouts
\usepackage{subcaption}
\usepackage[figuresleft]{rotating}
\usepackage{amsmath,amssymb,amsfonts}
\usepackage{bm}
\usepackage{paralist}
\usepackage[colorlinks=true,linkcolor=blue,urlcolor=blue,citecolor=blue,anchorcolor=red]{hyperref}
\usepackage{graphicx}
\usepackage{xcolor}
\usepackage{algpseudocode}
\usepackage{titlesec}
\usepackage{booktabs}
\usepackage{multirow}

\usepackage{xspace}
\newcommand{\btsp}{BioTac SP\xspace}

\newcommand{\ie}{\textit{i}.\textit{e}.}

\usepackage{todonotes} % for comments

\titlespacing*{\section}
{0pt}{1pt plus 2pt minus .2pt}{1pt plus .2pt}
\titlespacing*{\subsection}
{0pt}{1pt plus 2pt minus .2pt}{1pt plus .2pt}

\title{\LARGE \bf
GradTac: Spatio-Temporal Gradient Based Tactile Sensing}

\author{Kanishka Ganguly, Pavan Mantripragada, Chethan M. Parameshwara, \\
 Cornelia Ferm{\"u}ller, Nitin J. Sanket, Yiannis Aloimonos% <-this % stops a space

\thanks{All authors are associated with the Perception and Robotics Group, University of Maryland, College Park, MD, 20740.}%
\thanks{{\small{\texttt{\{kganguly, mppavan, cmparam9, fermulcm, nitinsan, jyaloimo\}@umd.edu}}}}%
}

\begin{document}

\maketitle
\thispagestyle{empty}
\pagestyle{empty}

\begin{abstract}
Tactile sensing for robotics is achieved through a variety of mechanisms, including magnetic, optical-tactile, and conductive fluid. Currently, the fluid-based sensors have struck the right balance of anthropomorphic sizes and shapes and accuracy of tactile response measurement. However, this design is plagued by a low Signal to Noise Ratio (SNR) due to the fluid based sensing mechanism ``damping'' the measurement values that are hard to model.

To this end, we present a spatio-temporal gradient representation on the data obtained from fluid-based tactile sensors, %This data representation 
which is inspired from neuromorphic principles of event based sensing.

We present a novel algorithm (GradTac) that converts discrete data points from spatial tactile sensors into spatio-temporal surfaces and tracks tactile contours across these surfaces. Processing the tactile data using the proposed spatio-temporal domain is robust, makes it less susceptible to the inherent noise from the fluid based sensors, and allows accurate tracking of regions of touch as compared to using the raw data.

We successfully evaluate and demonstrate the efficacy of GradTac on many real-world experiments performed using the Shadow Dexterous Hand, equipped with the \btsp sensors.
Specifically, we use it for tracking tactile input across the sensor's surface, measuring relative forces, detecting linear and rotational slip, and for edge tracking. We also release an accompanying task-agnostic dataset for the \btsp, which we hope will provide a resource to compare and quantify various novel approaches, and motivate further research.
\end{abstract}
\graphicspath{{\subfix{../images/}}}
\section{Introduction}\label{sec:intro}
Computational tactile sensing has myriad applications in robotics, especially in tasks related to grasping and manipulation. The robotics community has put a significant amount of effort into the design of hardware and algorithms to equip robots with tactile sensing capabilities that rival that of the human skin. Decades of research have led to the design of fluid based sensing mechanisms as the gold-standard for striking the balance between anthropomorphic shapes, sizes and responses. However, as computational algorithms have utilized such sensors widely, some largely unexplored issues still persist due to their non-linear behavior observed in both spatial and temporal responses due to external factors that are hard to model \cite{wettels2008biomimetic}.

% \chethan{I don't like this}
% In this work, we deal with one such state-of-the-art tactile sensing hardware, the \btsp tactile sensor, fitted on the Shadow Dexterous Hand. 

Primarily, these sensors have low Signal to Noise Ratios (SNR), owing to the use of a fluid-based transmission of forces from the skin to the sensing electronics which ``damps'' the values. Secondly, because of the non-uniform distribution of the sensing elements inside the mechanical construction,
%such as the \btsp, 
each sensing element has a different sensing range, and respective biases. 
These issues have prohibited the development of a standard representation of the data, and processing techniques have been designed engineered for a particular set of tasks rather than being general.

Many approaches have been proposed for interpreting the sensor data, with highly accurate computer models on one end \cite{narang2021interpreting, narang2021sim}, and a variety of signal processing techniques \cite{Wettels2011HapticFE} on the raw data on the other. Both these approaches are computationally expensive and need extensive hand-crafted calibration procedures for them to be operational.

On the contrary, biological systems calibrate for these environmental factors on-the-fly by processing tactile information as spikes or events, which provides advantages for transmission and processing along with built-in robustness. This ideology inspired neuromorphic engineers to develop sensors and low-power hardware \cite{brandli2014240}, that record and process events, as well as algorithms to compute events \cite{mitrokhin2018event, gallego2020event, sanket2020evdodgenet}. Recently event based hardware has become available for the research  community. The best known among these is a vision sensor called DVS \cite{brandli2014240, lichtsteiner2008128}, and another sensor is the event based audio cochlea \cite{yang20160}. Event-based processing has also been introduced to the olfactory domain \cite{jing2016signal} and for tactile data \cite{janotte2021touch}.

We propose a novel intermediate representation computed directly from the raw fluid-based tactile data such as that of the \btsp sensor. Instead of accurately simulating the deformations and forces on the sensor, as in \cite{narang2021sim,narang2021interpreting}, we compute robust features from the spatio-temporal changes in the tactile data, which carry essential information about the sensor's deformation and forces at the location of touch. The approach is computationally inexpensive and  sufficiently accurate to perform a series of tasks.

The main idea is to compute from a sequence of raw data, the significant changes in data values from individual sensors, which we call \textit{Tactile Events}, and then compute the essential tactile features from these events via a spatial  interpolation.
Specifically, by temporally accumulating the tactile events we construct surface contours, that can be used as a generic representation  for tracking touch across the \btsp skin. 
Our approach handles the challenges mentioned above, \ie, it can account for noise and individual sensor biases.
\graphicspath{{\subfix{../images/}}}

\subsection{Problem Formulation and Contribution}
The question we tackle in this work can be summarised as \textit{``What representation do we need to handle noisy data from a Fluid Based Tactile Sensor (FBTS)?''}. To answer this question, we draw inspiration from neuromorphic computing and propose a computational model for representing tactile data using spatio-temporal gradients. Our contributions are formally described next.

\begin{itemize}
    \item We present an intuition for the relationship between the volumetric deformations of the skin and fluid on a fluid based tactile sensor and spatio-temporal gradients. We further discuss why our method can robustly compute the maximal region of deformation. 
    
    \item We present a computational model to convert raw tactile signals from an FBTS into an interpolated spatio-temporal surface. This is then used to track regions of applied stimulus across the sensor's skin surface which corresponds to the regions of touch.
    
    \item We demonstrate the capabilities of our proposed approach on several real-world experiments, including detecting slippage during grasp, detecting relative direction of motion between fingers, and following planar shape contours.
    
    \item We  release a novel dataset containing the various experiments we perform on the \btsp. It can be used to validate not only our method, but also for comparing other tracking algorithms for the \btsp. and  help push the field forward.
\end{itemize}

\graphicspath{{\subfix{../images/}}}
\subsection{Prior Work}
Tactile sensors can broadly be categorized into two main categories: optical-tactile and direct-tactile sensing mechanisms.
The main tasks performed with tactile data found in the literature include: 1) estimation of the contact location and the net force vector, 2) estimation of high-density deformations on the sensor surface, 3) slip detection and classification, and 4) tracking object edges. We next discuss state-of-the-art works on using the various classes of tactile sensors and solving tasks related to those mentioned above.

\subsubsection{Signal Processing on Raw Data}
Studies that perform estimation directly on the sensor data include 
%Chia-Hsien \etal~in 
\cite{Lin2013PoC}, who present an analytical method to estimate the 3D point of contact and net force acting on the BioTac sensor based on electrode values, where they assume that electrodes measure force in the direction their normals.
%Su \etal~ 
\cite{schaal2015force} discuss several methods for force estimation from tactile data, including Locally Weighted Projection Regression and neural network based regression. They also present a signal processing technique for slip detection using the BioTac, comparing their results using an IMU.
%Sundaralingam \etal~
\cite{sundar2019force} introduce a method to infer forces from tactile data using a learning-based approach. They implement a 3D voxel grid to maintain spatial relations of the data, and use a convolutional neural network architecture to map forces to tactile signals.
\subsubsection{Physical Modelling Techniques}
Recently, some studies modeled a mapping between sensor readings and the field of deformations on the whole sensor surface. 
%Narang, \etal 
\cite{narang2021interpreting} presented a finite element model for the 19 taxel BioTac sensor and demonstrated the most accurate simulations of the sensor thus far. They relate forces applied to specific locations to the sensor's skin deformation. They learn using  data they collected, the mapping between 3D contact locations and netforce vectors to the 19 taxel readings, and then by combining the FEM simulation and experimental data they extrapoloate  a mapping between taxel sensor measurements and skin deformations and  vice-versa.
In \cite{narang2021sim} the authors extended this work  using variational autoencoder networks to represent both FEM deformations and electrode signals as low-dimensional latent variables, and they performed cross-modal learning over these latent variables. This enhanced the accuracy  of the mapping between taxel readings and skin deformations previously  obtained. However, they also showed that for unseen indenter shapes these methods poorly generalise in predicting deformation magnitudes and distributions from electrode values.
\subsubsection{Approaches Using TacTip Sensor}
%Lepora \etal in~
\cite{lepora2019tactipcontours} using a TacTip optical-tactile sensor (\cite{wardcherrier2018tactip}) learn via a CNN to perform reliable edge detection, and then use that in a visual servoing control policy for tracking and moving across object contours.
In related work by the authors  (\cite{cramphorn2018tactipvoronoi}), they present a Voronoi tesselation based processing pipeline  to predict contact location, as well as shear direction and magnitude on the surface of the sensor. This method is novel in that it does not use any classification or regression techniques and is purely analytical in nature.
\subsubsection{Event-Based Sensors}
%Taunyazov \etal in
\cite{taunyazov20event} use the NeuTouch, a novel event-based tactile sensor along with a Visual-Tactile Spiking Neural Network to perform object classification and rotational slip detection. They also perform ablation studies with an event-based visual camera, and compare their spiking neural networks to traditional network architectures like 3D convolutional networks, and Gated Recurrent Units.
\subsubsection{Inspiration from Prior Work}
We use the prior work described above as a source of motivation for our pipeline, and we attempt to use the validated experiments in them as a proof of concept of our approach. We perform slip detection experiments as in 
%Su \etal~ 
\cite{schaal2015force}, perform edge tracking using visual servoing as in 
%Lepora \etal in~
\cite{lepora2019tactipcontours} and compute forces from touch as described by %Sundaralingam \etal~
\cite{sundar2019force}.
\graphicspath{{\subfix{../images/}}}

\subsection{Organization of the paper}
Our paper is organized as follows:
In Sec.~\ref{sec:method}, we present the motivation for using the \btsp sensor for tactile sensing, and how our method is a practical solution to the challenges posed by this particular type of sensor. We describe in detail why the spatio-temporal gradients are an intuitive way for computing  features of deformation on the \btsp.

Sec.~\ref{sec:approach} discusses our high-level pipeline, and our experimental setup. We then go into detail regarding our algorithm to generate spatio-temporal gradients, \ie~events from raw tactile data, and then discuss how we generate contour surfaces from these events. Lastly, in this section we discuss how we use these spatio-temporal surfaces to track touch stimulus across the \btsp skin.

In Sec.~\ref{sec:experiments} we demonstrate our pipeline on three distinctly different tasks, and discuss their results and outputs. We first show that our contour surfaces are able to accurately track tactile stimulus in motion across the surface of the \btsp skin. We then discuss results on experiments involving varying applied forces on the \btsp, where we show that our contour surfaces can distinguish between various levels of force.
Lastly, we employ our algorithm on a more practical task of slippage detection during grasping, where we detect time of slippage, and distinguish between longitudinal and rotational slippage types.\\
\textit{All images presented here are best viewed on a color display at 200\% zoom.}
\graphicspath{{\subfix{../images/}}}
\section{Method}
\label{sec:method}
\subsection{Motivation}
\label{subsec:motivation}
% \chethan{Seems out of place, maybe put this in experiments}
We consider for our work the \btsp tactile sensor, which comes with a unique sensing mechanism as compared to other contemporary tactile sensors. Tactile sensing mechanisms, as they are available commercially today, lie on a spectrum ranging from \textit{accurate sensing capabilities} on one side to \textit{biomimetic form-factors} on the other. Most sensors on this spectrum make trade-offs on form factor to provide high accuracy. The \btsp is one particular sensor that strikes a right balance and is in the middle of the range, where we have a physical shape and sensing mechanism very close to the human finger tip, but this comes at the cost of accuracy and fidelity of sensing. 

\subsection{Challenges with Fluid-Conductive Sensors}
\label{subsec:challenges}
Unlike optical-tactile, magnetic or capacitive tactile sensors, fluid based tactile sensors use a conductive fluid to transmit electrical impulses from spatially distributed excitation electrodes to a few sensing locations (taxels) distributed over a solid core. The values generated by the taxels are thus primarily dependent on the characteristics of the fluid, specifically its conductivity.\\
The conductivity of a fluid, such as the electrolytic solution present in the \btsp sensor is non-linearly related to various external factors. These include, but are not limited to the temperature of the fluid, the humidity of the surroundings, the area and distance between the excitation and sensing electrodes, and the concentration of the conductive fluid. Each of these factors contribute non-linearly (\cite{wettels2008biomimetic}) to the  noise of the individual  taxels. Furthermore, the noise characteristics of the sensor electronics are also non-linear, which further exacerbates the situation.\\
We also need to consider sources of noise in the electronic implementation of each taxel's sensing mechanism, which include amplification and analog-to-digital conversion circuitry among others.

\subsection{Modelling Fluid-Conductive Sensors}
\label{subsec:modelling}
While it might be feasible to model each of the aforementioned sources of noise independently and in isolation, the combination and interactions between them when considered together in the system makes it an arduous task. There have been several attempts to develop a physical model of the BioTac sensor, the most recent of which is presented in the work by~\cite{narang2021sim}. In this, the authors present a finite element model (FEM) of an ideal BioTac sensor, and provide an accurate simulation of the skin, the sensing core, and the internal fluid. While the FEM approach provides a physically accurate measurement of the deformation of the skin and fluid based on force stimuli, it does not account for the sources of noise described earlier. This is because the model of the sensor electronics is not considered along with the computational challenges of fluid modelling. Currently, to the best of our knowledge, there is no mathematical model between sensor readings and skin deformations, thereby inhibiting research in this area when utilizing raw sensor measurements.
% \nitin{We need to make a strong point of how we can avoid this}

\subsection{Bio-inspired motivation for logarithmic change}
\label{subsec:log-change}
In our work, we draw inspiration from nature regarding 
%, developing an ``inuition'' based on 
how changes over the skin surfaces may be  related to location of touch and relative forces. To this end, we break away from the core robotics ideology that one requires a complex and accurate mathematical model or a very high quality sensor to perform useful tasks. In particular, we are driven by nature's efficient and parsimonious implementations which perform amazingly well with minimal quality sensors and very simple computing. 

To build such an efficient data representation for fluid based sensors, we look at the Weber-Fechner Laws, which state that the perceived stimulus on any of the human senses is related via an exponential function to the actual stimulus. As a result, humans perceive stimuli such as touch, sound or light as the changes in the logarithm between existing values and new ones.
It is thus not surprising that the manufacturers of the \btsp sensor, who designed it to be as anthropomorphic as possible, also recommend that the best way to process the data from such fluid-based sensors is to use relative changes instead of raw taxel values.

In practice, the two main challenges with the \btsp sensor are that
\begin{inparaenum}
  \item the different taxels do not have same baseline value, and
  \item the taxel values exhibit a low signal to noise ratio.
\end{inparaenum}

By computing only the taxel changes on a logarithmic scale, our values become independent of the baseline and are more robust to noise, thus tackling both aforementioned issues.

\subsection{Computing events from raw data}
\label{subsec:computing-events}
We consider two consecutive packets of taxel data, at times $t$ and $t+\delta$ respectively. Each of these packets contains the raw taxel values $X^{t}_{k=1\dots24}$ and $X^{t+\delta}_{k=1\dots24}$. We then compute the logarithmic change between each of the $j \in {1, 24}$  consecutive taxels, and fire an event when  the logarithm of the value at a taxel increases or decreases by a threshold value $\tau$. That is, when: 
\begin{equation}
| \ln X^{t= i + \delta}_{j} - \ln X^{t= i}_{j} | > \tau
\end{equation}
In other words, a positive event is said to be ``fired'' when 
\begin{equation}
\ln X^{t= i + \delta}_{j} > e^{\tau}\ln X^{t= i}_{j}
\end{equation}
and a negative event when 
\begin{equation}
\ln X^{t= i}_{j} > e^{\tau}\ln X^{t= i+ \delta}_{j}
\end{equation}
This gives us intermediate taxel values between times $t$ and $t +\delta$, and their respective timestamps.

We know from the design of the sensor, as well as ideal sensor simulations that the largest change in the values of the taxels correlates with the region of highest tactile stimulus. Also, this change is dependent on the forces already present on the region of touch, and reaches saturation and demonstrates hysteresis in the raw values. Our algorithm accounts for that by non-linearly interpolating the taxel data, on the log scale.
The previously obtained taxel events thus give us a temporal gradient over the change in taxel values, caused by the deformation of the skin and fluid because of the applied force stimulus. Intuitively, these intermediate events between two discrete taxel data values signify change in localized volume of the skin and fluid over time due to the applied tactile stimulus.

As part of our algorithm, we then process these spatially discrete events for each of the 24 taxel locations and convert them into a continuous, interpolated surface. We use a Voronoi tesselation of the discrete and irregular grid, and perform Natural Neighbors Interpolation to construct an event surface, that gives us an interpolated event value at each point. This surface indirectly depicts the deformation of the skin and fluid, due to applied stimulus. Since the deformation due to applied forces on the \btsp is greatest at the region of touch, we generate iso-contours of the event surface, and consider only the maximal valued contour as the region of touch.
\graphicspath{{\subfix{../images/}}}

\section{Our Approach}\label{sec:approach}

\subsection{Pipeline}
\begin{figure*}
    \centering
    \includegraphics[width=\linewidth]{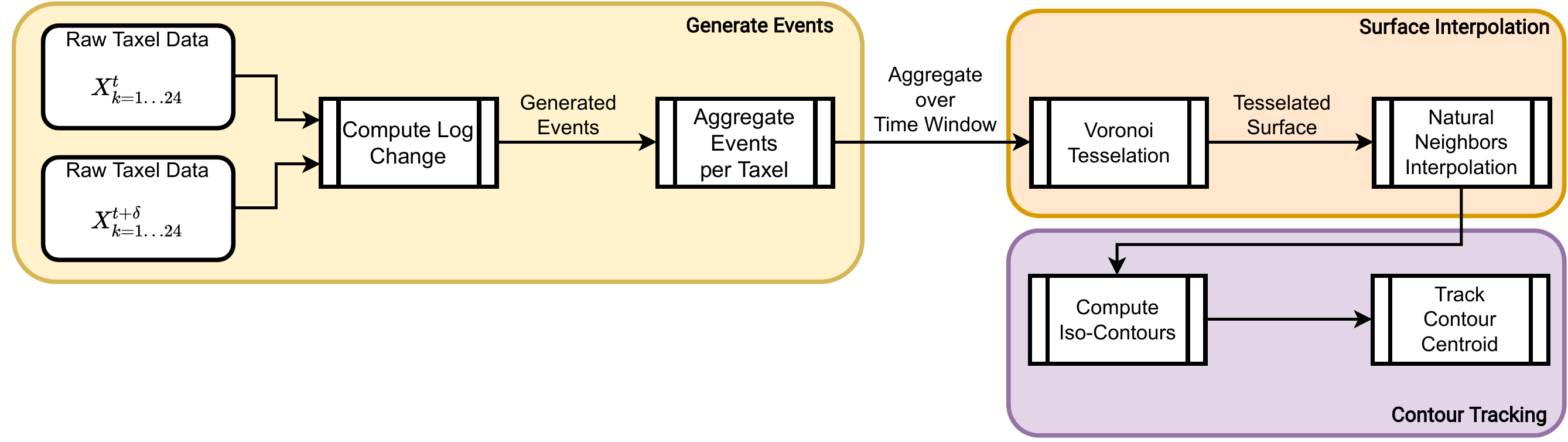}
    \caption{High-Level Organization of Our Pipeline}
    \label{fig:highlevel}
\end{figure*}
Fig.~\ref{fig:highlevel} shows an overview of the proposed framework, where we start with 24 points of raw tactile data from the BioTac SP sensors and generate a contact trajectory as output.
The pipeline involves converting the raw data into events, aggregating said events by spatial clusters, performing Voronoi Tessellation on the aggregate events, and then using the interpolated values to generate a contour plot whose centroid is tracked over time. We elaborate each of the steps of our pipeline below.

\subsection{Setup and Methodology}

Our hardware setup consists of a UR-10 manipulator equipped with the Shadow Dexterous Hand, which has the \btsp sensors attached to each finger tip.
The \btsp provides a ROS interface to obtain the raw data, at a rate of 100 Hz. This data consists of 24 electrode values which we term ``taxels'' (tactile element), as well as overall fluid pressure and temperature flux. For our pipeline, we use only the 24 taxel readings. These readings are the result of forces due to contact and the resultant compression of the skin and the enclosed fluid.
The nature of our pipeline allows for processing readings from any other tactile sensor, as long as they are spatially distributed across some surface, and  timestamps for each data packet are provided. We perform basic min-max normalization and Savitzky-Golay filtering before using the data.
Our event-generation algorithm is influenced by  principles of event-based sensors, which record logarithmic changes of signal on individual sensing elements, independently and asynchronously.

%, so it is worthwhile to explain the fundamentals behind that before moving on to event-based tactile processing.
%Event-based vision sensors have, unlike traditional CMOS vision sensors, independently responsive pixels across the entire sensor. This means that when light reaches the sensor, each pixel is triggered independent of its neighbors, based on a predetermined threshold which compares the previous light intensity value to the current.

\subsection{Generating Events from Raw Tactile Data}\label{subsec:tactile_gen}

In Sec.~\ref{subsec:motivation}, we established that our approach does not approximate the entire sensor's surface but only the regions with maximum tactile stimulus. 
%We are interested in computing the change in volume deformation of the surface at relatively small intervals of time, at the region of applied tactile force.
The data from the \btsp sensor is obtained at a rate of 100 Hz, or one packet of data every 0.01 seconds. Our method computes the number of events at each taxel, where each event corresponds to the change of some threshold value $\tau$.
%, and changes of multiple $\tau$'s corresponding to multiple events.
%However, even between two consecutive data points, there are many infinitesimal ``steps'' that the surface took to get from one deformed state to the next. Our event-generation algorithm simulates these intermediate deformations by looking at change of values from time $t$ to $t + \delta$ and dividing them up into smaller ``events'' that are fired based on some threshold $\tau$.
This essentially decides the granularity of change we are interested in measuring, and more events correspond to larger change, which is correlated to the amount of force that was applied to a particular region. For each event triggered, we also generate a corresponding timestamp between $t$ and $t +\delta$.
Taking inspiration from the Weber-Fechner laws of psychophysics mentioned earlier in Sec.~\ref{sec:intro}, we trigger events based on the natural log of the threshold $\tau$. This intuitively means that the frequency of events are higher initially at time $t$, and gradually taper off as we get closer to the value at time $t+\delta$.

Once the events have been computed for all the raw tactile data points, we aggregate them into temporal frames.
%with an adjustable window size. 
The size of the temporal window used for aggregation is an important heuristic that can be fine tuned to favor robustness to noise or allow for a more sensitive tactile response. %Here, for each of the 24 taxels, we get one spatio-temporal frame that contains the aggregate event count.

\subsection{Natural Neighbors Based Interpolation}
\label{subsec:natural_neighbors_interpolation}

\begin{figure}
    \begin{minipage}[t]{0.2\textwidth}
        \centering
        \includegraphics[width=1.0\linewidth]{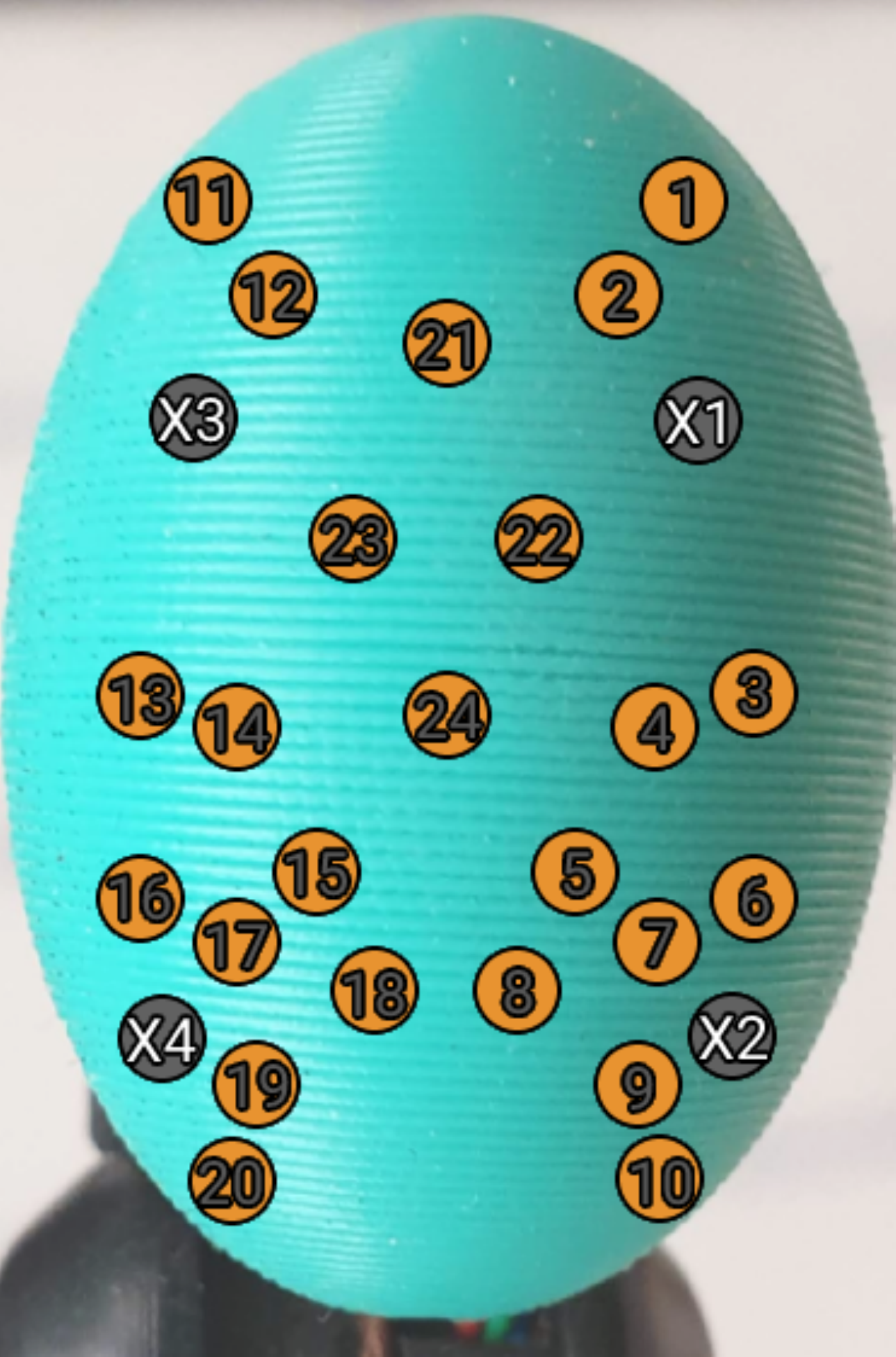}
        (A)
    \end{minipage}\hfill%
    \begin{minipage}[t]{0.2\textwidth}
        \centering
        \includegraphics[width=1.0\linewidth]{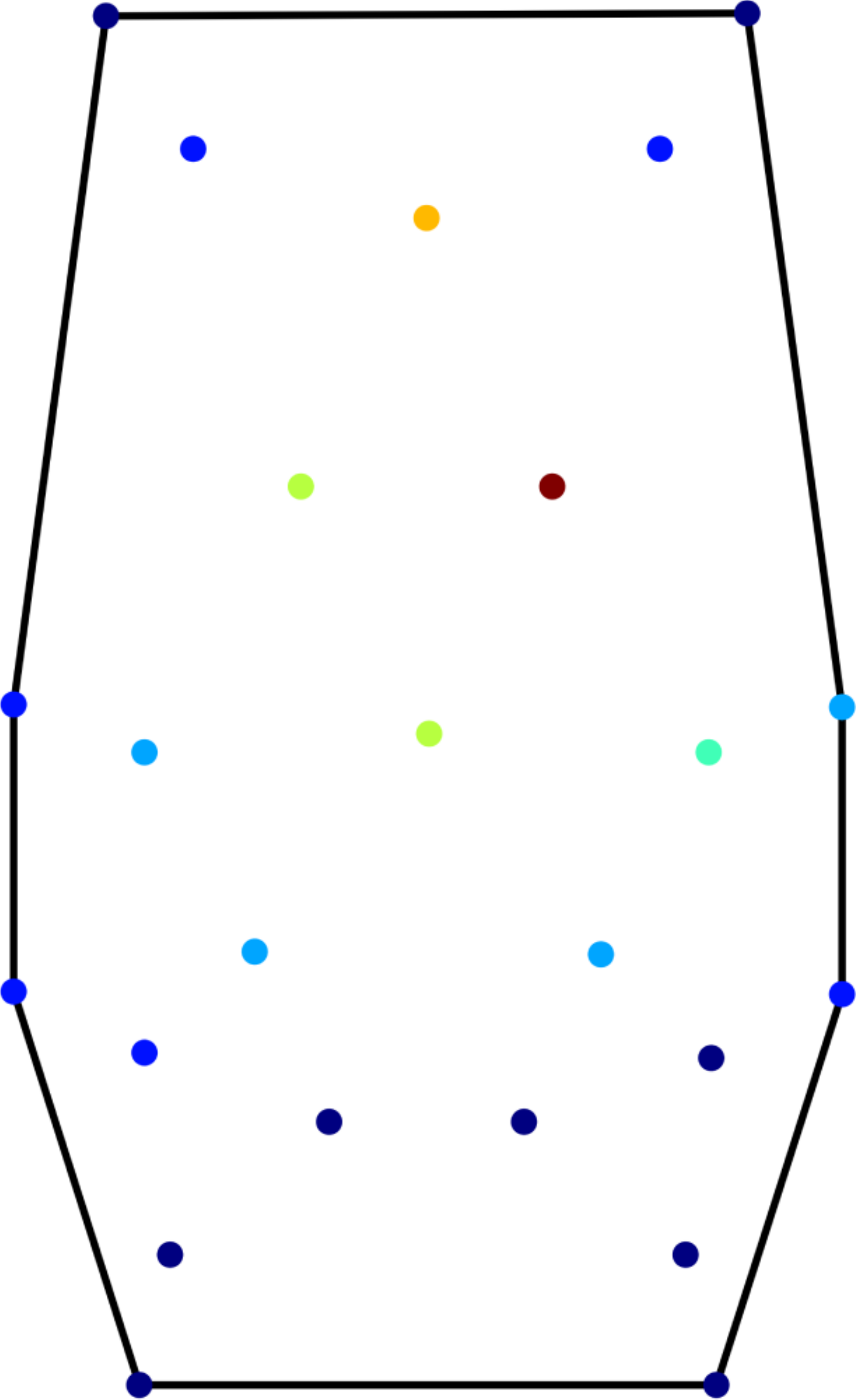}
        (B)
    \end{minipage}\hfill%
    \begin{minipage}[t]{0.2\textwidth}
        \centering
        \includegraphics[width=1.0\linewidth]{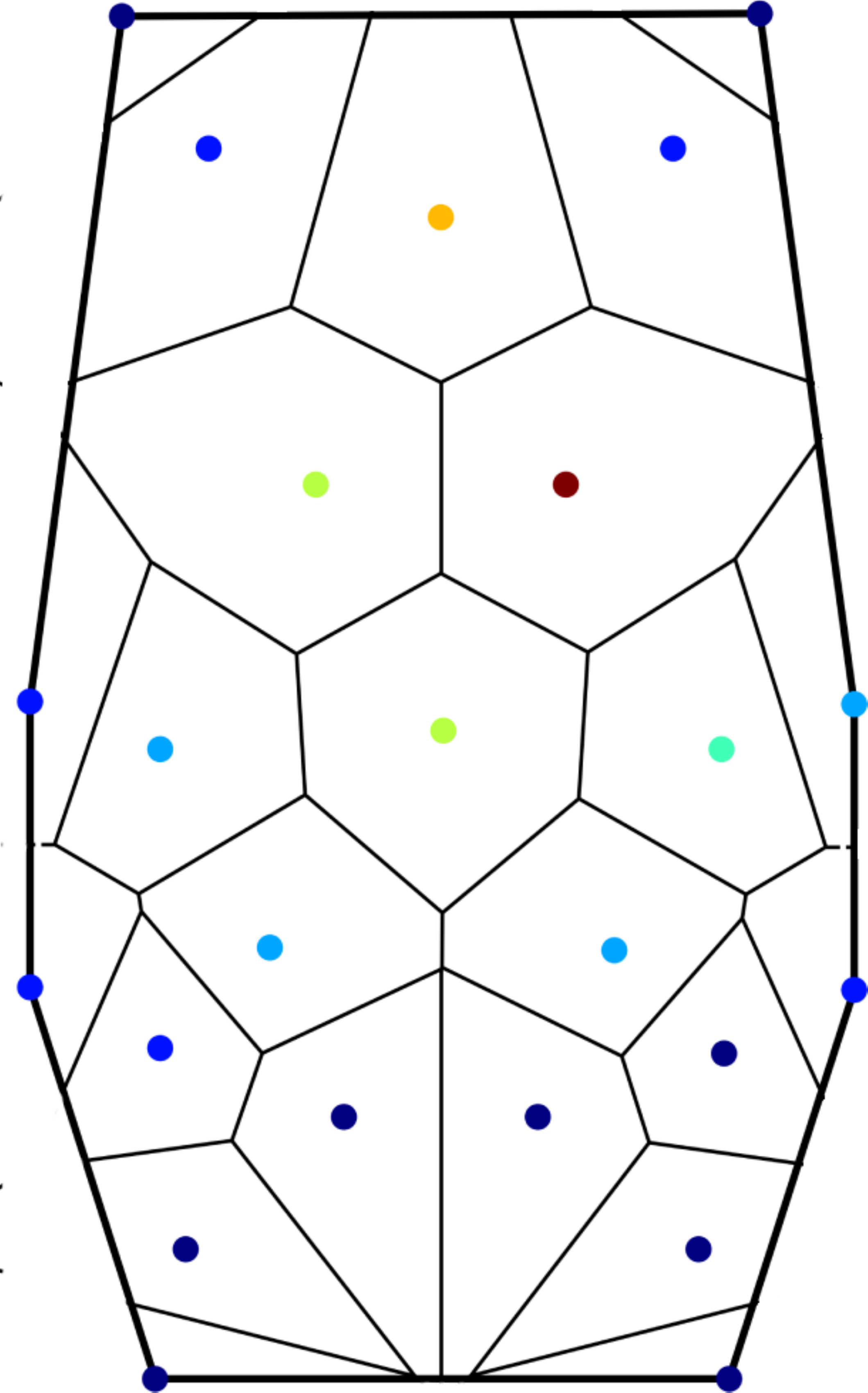}
        (C)
    \end{minipage}\hfill%
    \begin{minipage}[t]{0.2\textwidth}
        \centering
        \includegraphics[width=1.0\linewidth]{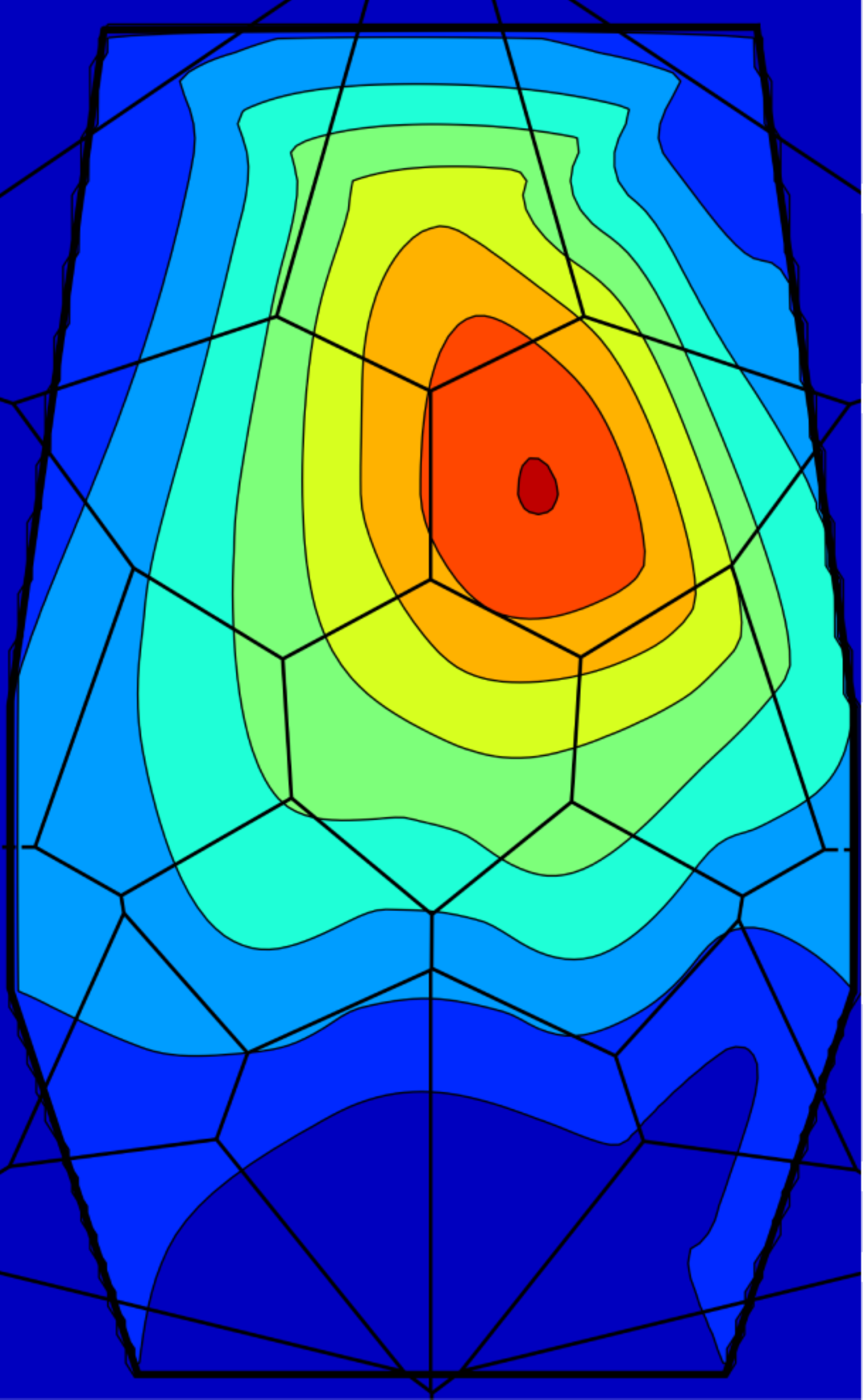}
        (D)
    \end{minipage}
    \caption{Contour Generation Pipeline. (A) 24 taxel locations, projected onto 2D plane, (B) Initial event aggregates per taxel, (C) Voronoi tesselation of the grid, (D) Contours generated from the interpolated surface}
    \label{fig:contour_generation_pipeline}
\end{figure}

The 24 taxels are located in some 3D space inside the \btsp, as per the sensor's design. We project these ellipsoidal locations onto a 2D surface, shown in Fig.~\ref{fig:contour_generation_pipeline}(A), to get an irregular grid of locations on a plane. For each of these 24 2D points, we have the aggregate event counts, as shown in Fig.~\ref{fig:contour_generation_pipeline}(B).

We proceed to perform Voronoi tessellation of this grid, based on the aggregate event values, shown in Fig.~\ref{fig:contour_generation_pipeline}(C). Compared to other methods of interpolation, such as Inverse Distance Weighting or Gaussian interpolation, Voronoi tessellation provides a more accurate representation of the underlying function we are trying to interpolate.
Considering the unstructured nature of our data, i.e. an irregular grid of taxels, traditional methods of interpolation do not take into account the different areas of influence of each taxel when computing the interpolated function. Voronoi tesselation partitions the space proportional to the ``strength'' of each sample point, by ``stealing'' some area from the neighboring polygons any time a new point is interpolated \cite{voronoi2021}.

This is mathematically represented by:
\begin{equation}
    G(\mathbf{x}) = \sum_{i=1}^{n} w_i(\mathbf{x})f(\mathbf{x}_i)
\end{equation}
where $G(\mathbf{x})$ is the estimate computed at $\mathbf{x}$, and $w_i$ are weights, and $f(\mathbf{x}_i)$ are the known data values at $\mathbf{x}_i$, which are obtained from the 24 event aggregate values.

Intuitively, Voronoi tessellation partitions the space into irregularly shaped polygons that are proportional to (or representative of) the ``force'' exerted on each taxel location. This is better than say, nearest neighbors interpolation which interpolates force values uniformly around each taxel.

The results of the Voronoi tessellation are used to interpolate points on the 2D surface of the \btsp, resulting in a continuous surface (Fig.~\ref{fig:contour_generation_pipeline}(D)) whose values correspond to the amount of force on  each taxel, and consequently, the deformation of that region of the \btsp skin.

\graphicspath{{\subfix{../images/}}}

\section{Dataset}
There is a lack of standardized datasets in the tactile sensing community, especially when sensors like the \btsp are concerned. Most datasets available today are task-specific, or are from optical-tactile sensors. This makes quantitative comparisons difficult for novel algorithms being introduced to the field.

As part of our work, we are releasing an accompanying dataset  on tactile motion on the \btsp sensor, which is independent of any particular task. The dataset samples %can broadly be classified into the following categories:
include the following:
\begin{itemize}
    \item Tactile responses from various indenter sizes, applied at different forces
    \item Motion across the sensor surface in various directional trajectories. We include 
        \begin{inparaenum}[a)]
        \item top-to-bottom,
        \item bottom-to-top,
        \item left-to-right,
        \item right-to-left,
        \item diagonal top-to-bottom,
        \item diagonal bottom-to-top,
        \item clockwise and
        \item counter-clockwise
        \end{inparaenum}data samples.
    \item Longitudinal slippage for various objects from a labelled list of objects, as well as the ground-truth timestamps for slip events.
    \item Rotational slippage for cylindrical object on a constant-speed turntable, as well as the ground-truth timestamps for slip events.
\end{itemize}

All our data is presented in both NumPy and CSV data formats, and includes all raw 24 taxel values as well as their timestamps. For ease of adoption and use, we eschew the ROS Bag format in this dataset, but it may be made available on request.
\graphicspath{{\subfix{../images/}}}

\section{Experiments and Results}\label{sec:experiments}

\subsection{Experimental Setup}\label{subsec:experimental_setup}
The hardware used to perform all experiments consists of a UR-10 manipulator equipped with a Shadow Dexterous Hand, with one \btsp sensor attached to each of the five finger tips.\\
Alongside the 24 taxel values from each \btsp sensor, the setup also provides us with the 6 DoF pose of the arm and each finger, relative to a world coordinate system at the base of the manipulator. This information is used in the shape tracking experiments.
\graphicspath{{\subfix{../images/}}}

\subsection{Tracking Location of Touch}\label{subsec:contact_location}

\begin{figure*}
    \begin{minipage}[t]{0.2\textwidth}
        \centering
        \includegraphics[width=1.0\linewidth]{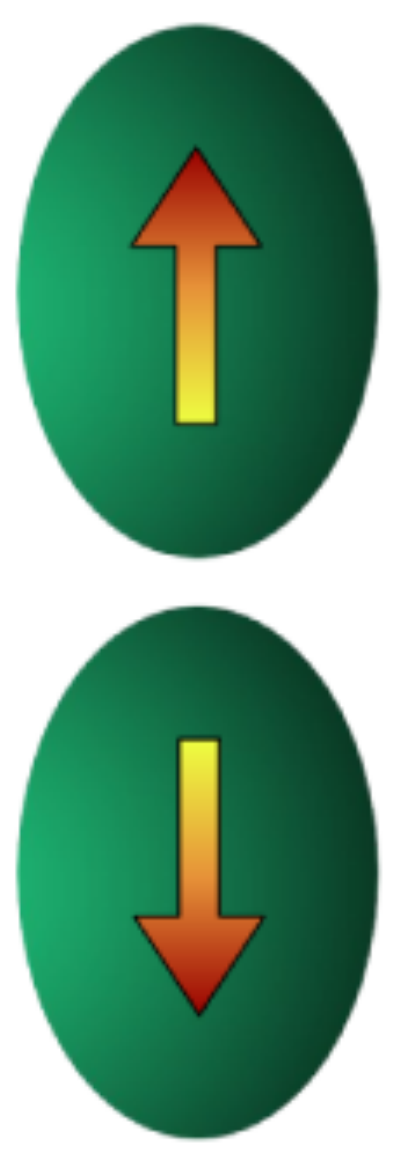}
        (A)
    \end{minipage}\hfill%
    \begin{minipage}[t]{0.2\textwidth}
        \centering
        \includegraphics[width=1.0\linewidth]{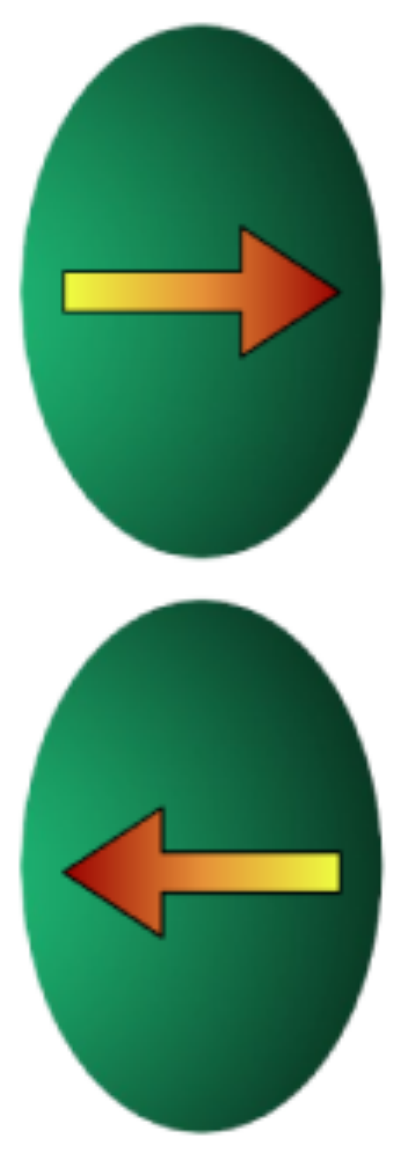}
        (B)
    \end{minipage}\hfill%
    \begin{minipage}[t]{0.2\textwidth}
        \centering
        \includegraphics[width=1.0\linewidth]{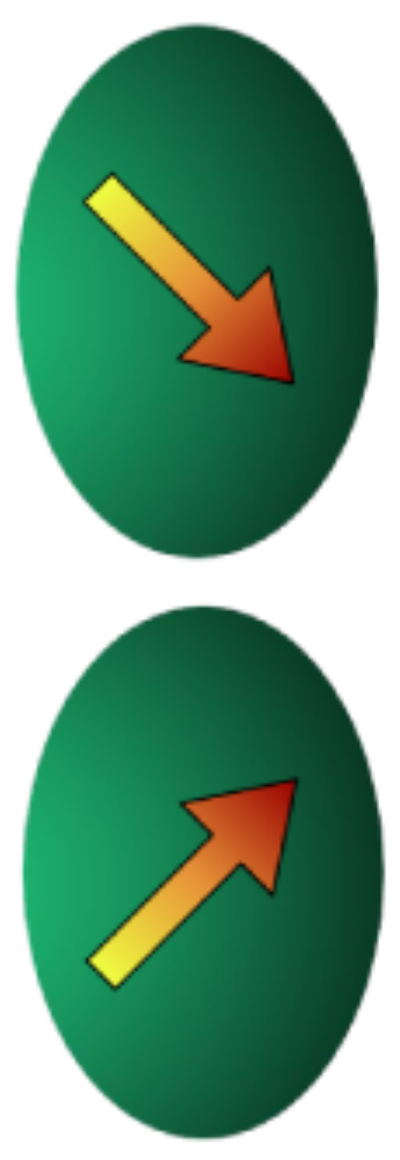}
        (C)
    \end{minipage}\hfill%
    \begin{minipage}[t]{0.2\textwidth}
        \centering
        \includegraphics[width=1.0\linewidth]{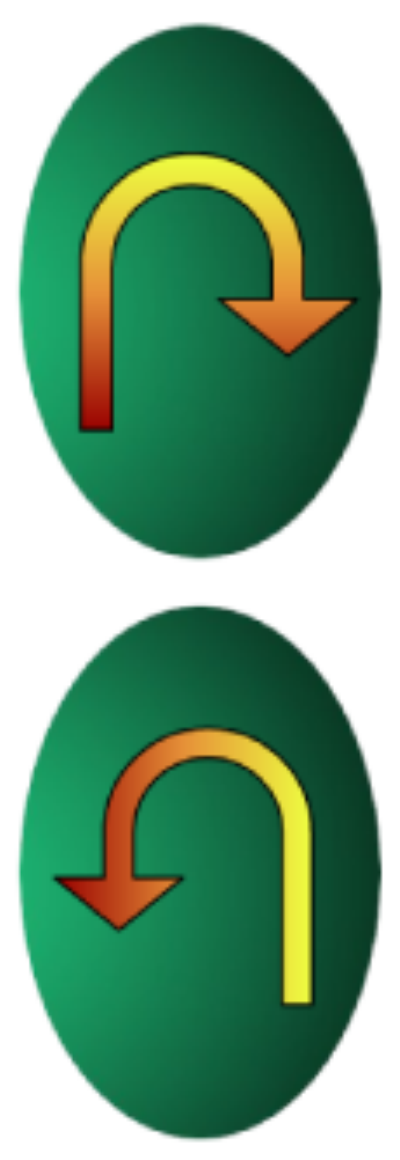}
        (D)
    \end{minipage}
    \caption{Touch Tracking Trajectories. (A) Up and Down Trajectories, (B) Left and Right Trajectories, (C) Diagonal Trajectories, (D) Circular Trajectories}
    \label{fig:touch_tracking}
\end{figure*}

\begin{figure*}
    \centering
    \begin{minipage}[t]{0.45\linewidth}
        \centering
        \includegraphics[width=1.0\linewidth]{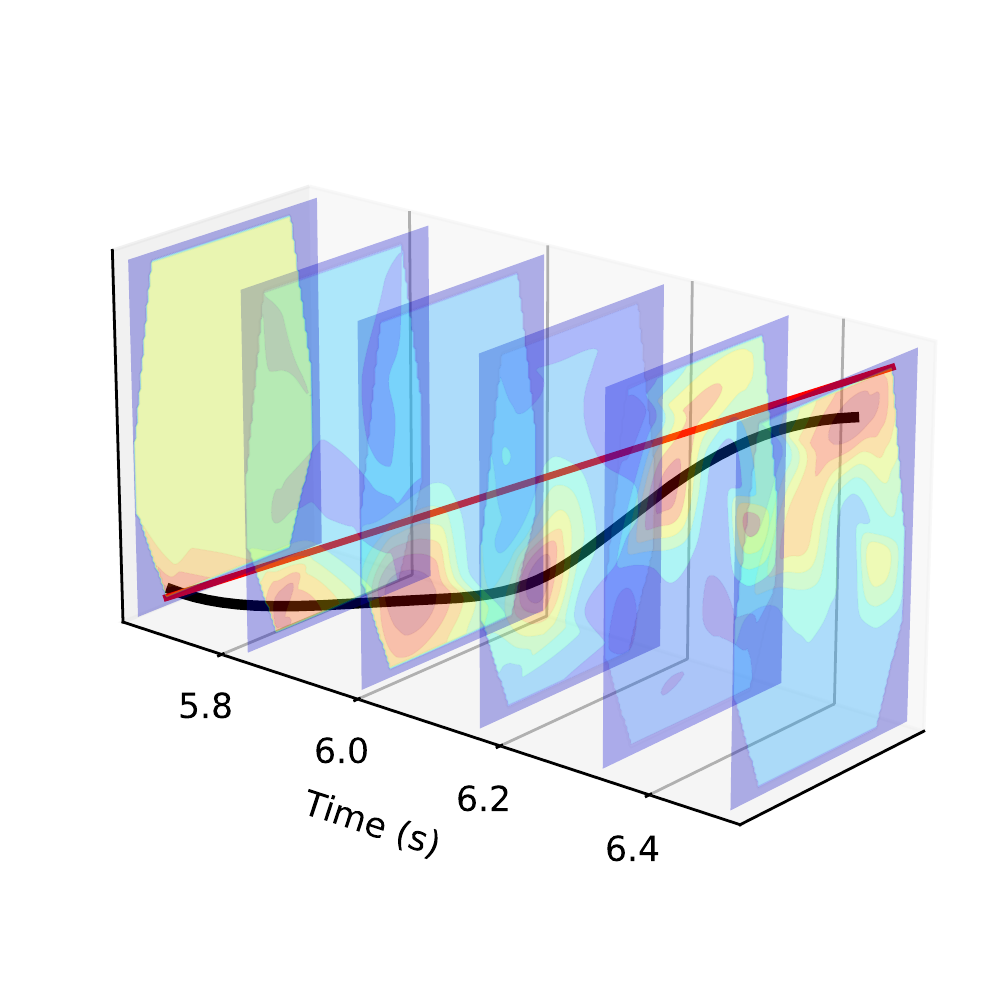}
        (A)
    \end{minipage}\hfill%
    \begin{minipage}[t]{0.45\linewidth}
        \centering
        \includegraphics[width=1.0\linewidth]{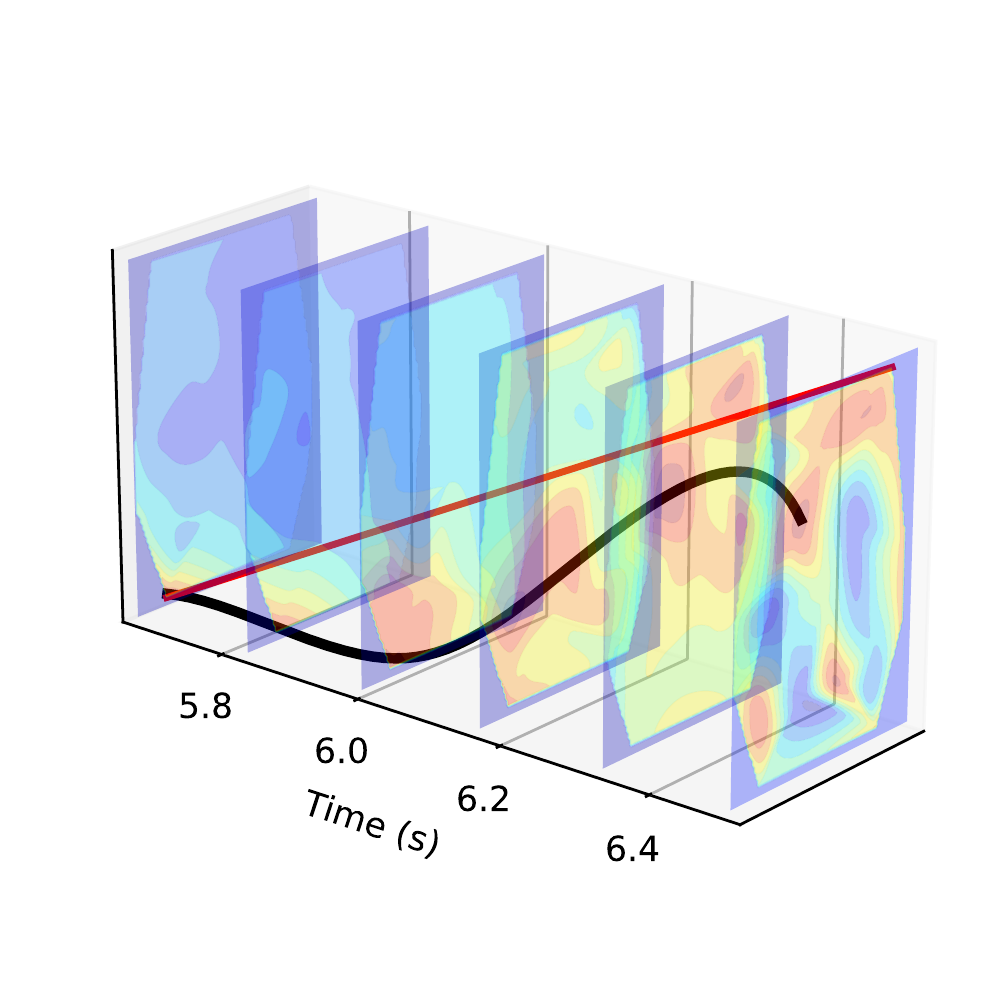}
        (B)
    \end{minipage}\\
    \begin{minipage}[t]{0.45\linewidth}
        \centering
        \includegraphics[width=1.0\linewidth]{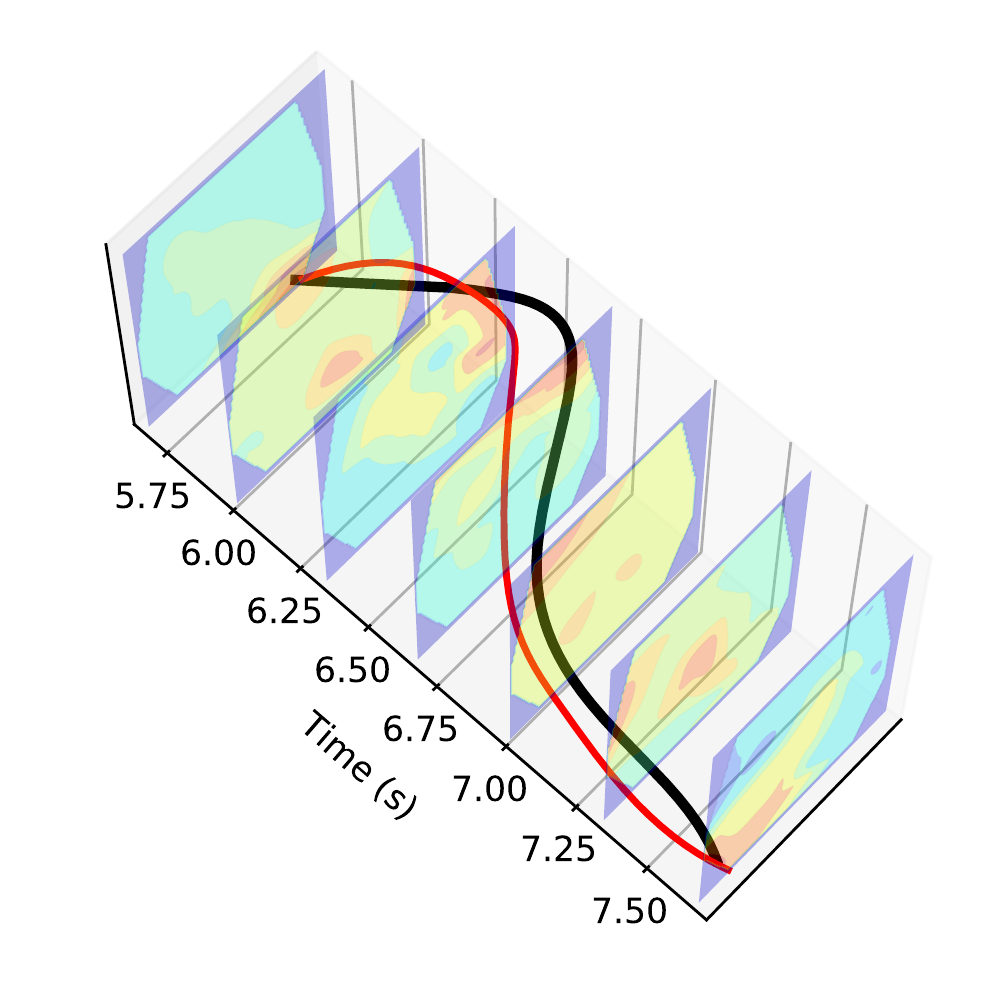}
        (C)
    \end{minipage}\hfill%
    \begin{minipage}[t]{0.45\linewidth}
        \centering
        \includegraphics[width=1.0\linewidth]{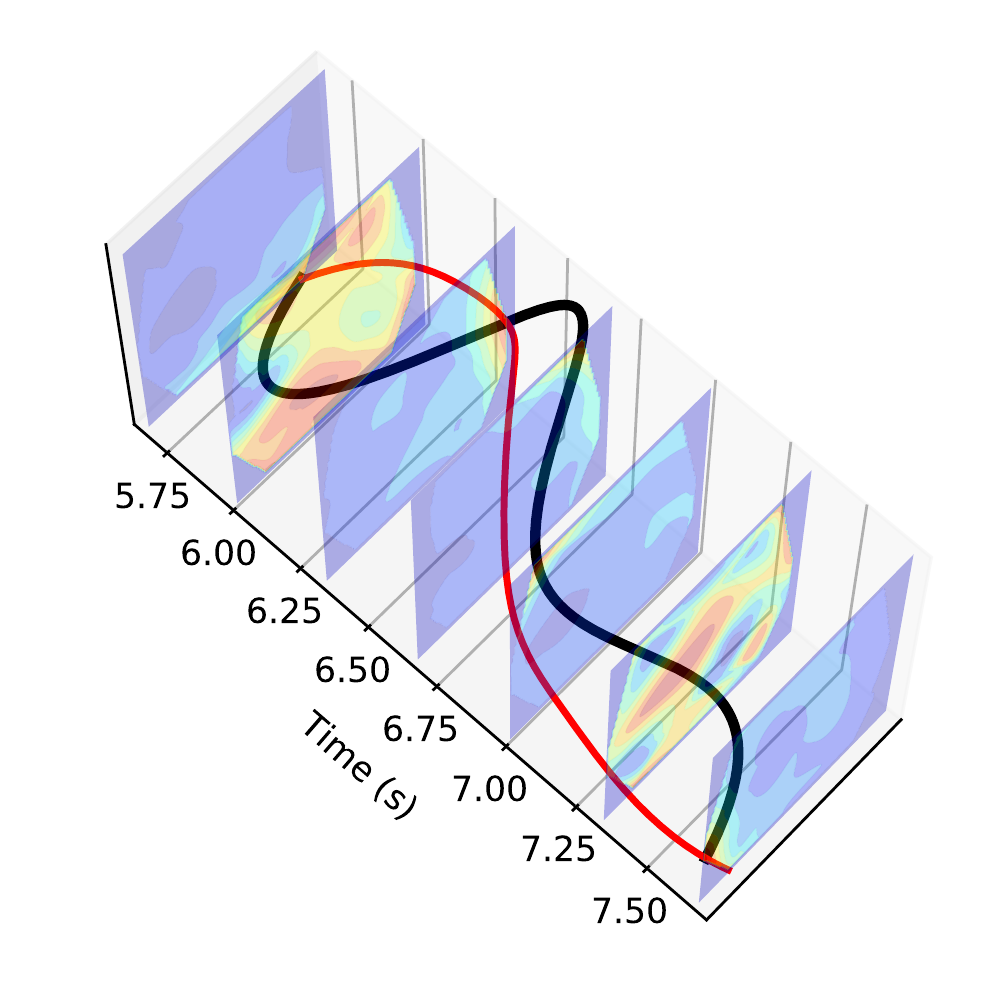}
        (D)
    \end{minipage}
    \caption{Touch Tracking Trajectory Plots. The red line denotes the ground truth trajectory.\\
    (A) Diagonal Trajectory using Events Data, (B) Diagonal Trajectory using Raw Data, (C) Circular Trajectory using Events Data, (D) Circular Trajectory using Raw Data}
    \label{fig:touch_tracking_plots}
\end{figure*}

\begin{figure*}[ht!]
    \begin{minipage}[t]{0.2\textwidth}
        \centering
        \includegraphics[width=1.0\linewidth]{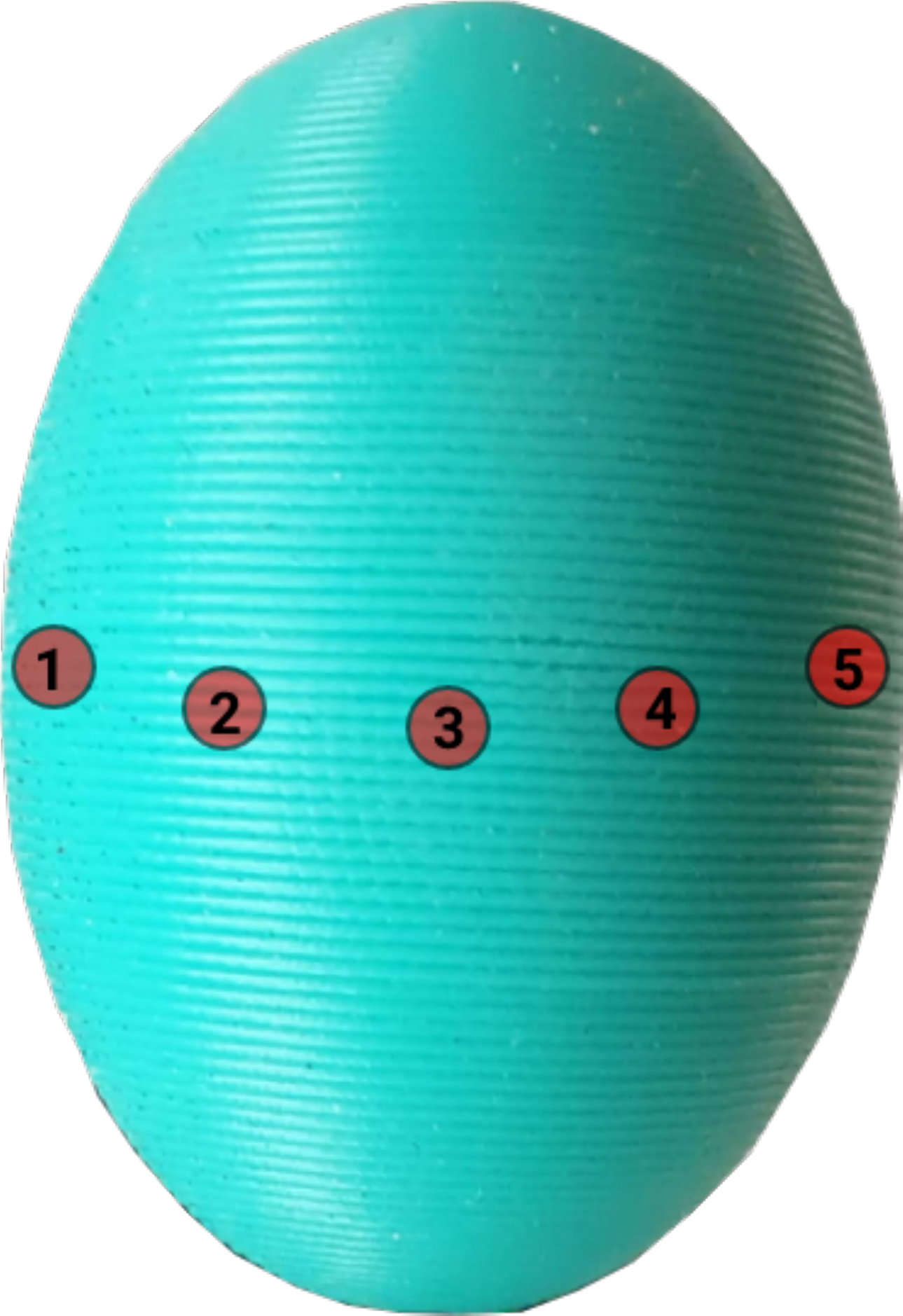}
        (A)
    \end{minipage}\hfill%
    \begin{minipage}[t]{0.2\textwidth}
        \centering
        \includegraphics[width=1.0\linewidth]{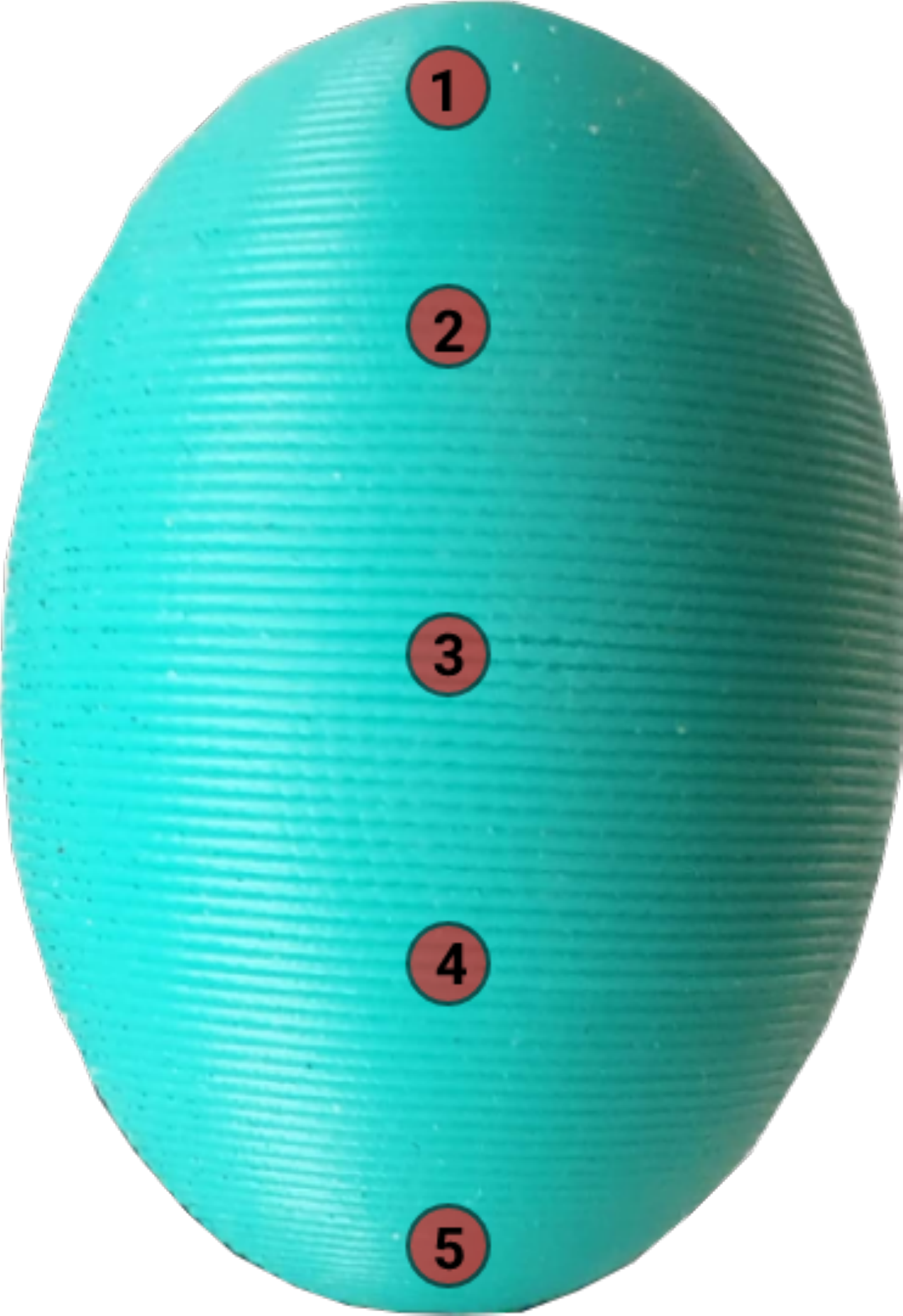}
        (B)
    \end{minipage}\hfill%
    \begin{minipage}[t]{0.2\textwidth}
        \centering
        \includegraphics[width=1.0\linewidth]{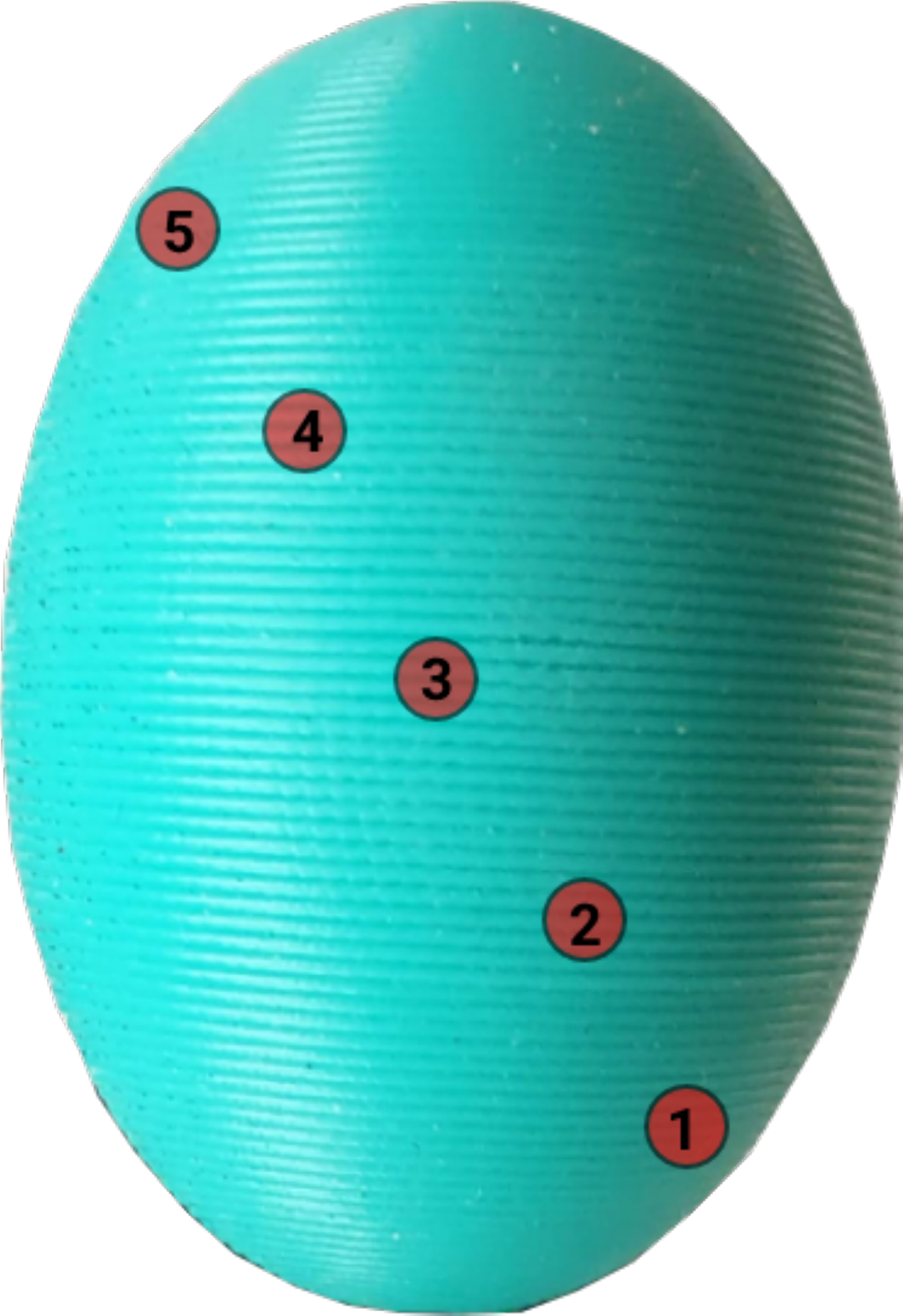}
        (C)
    \end{minipage}\hfill%
    \begin{minipage}[t]{0.2\textwidth}
        \centering
        \includegraphics[width=1.0\linewidth]{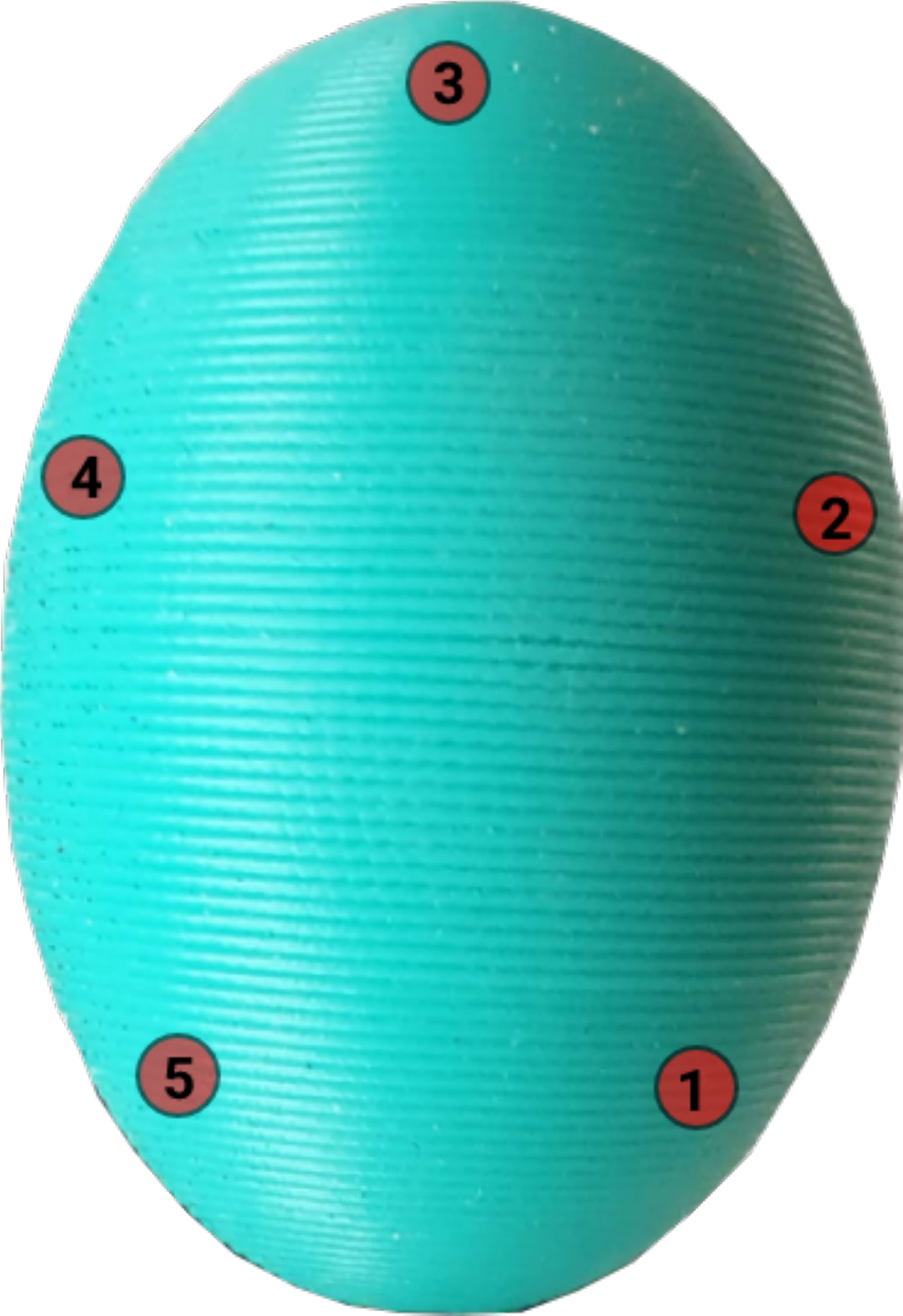}
        (D)
    \end{minipage}
    \caption{Touch Tracking Ground Truth Marker Locations. (A) Markers for tracking horizontal trajectory, (B) Markers for tracking vertical trajectory, (C) Markers for tracking diagonal trajectory, and (D) Markers for tracking circular trajectory}
    \label{fig:touch_tracking_markers}
\end{figure*}

\begin{figure*}
    \begin{minipage}[b]{0.5\textwidth}
        \centering
        \includegraphics[width=1.0\linewidth]{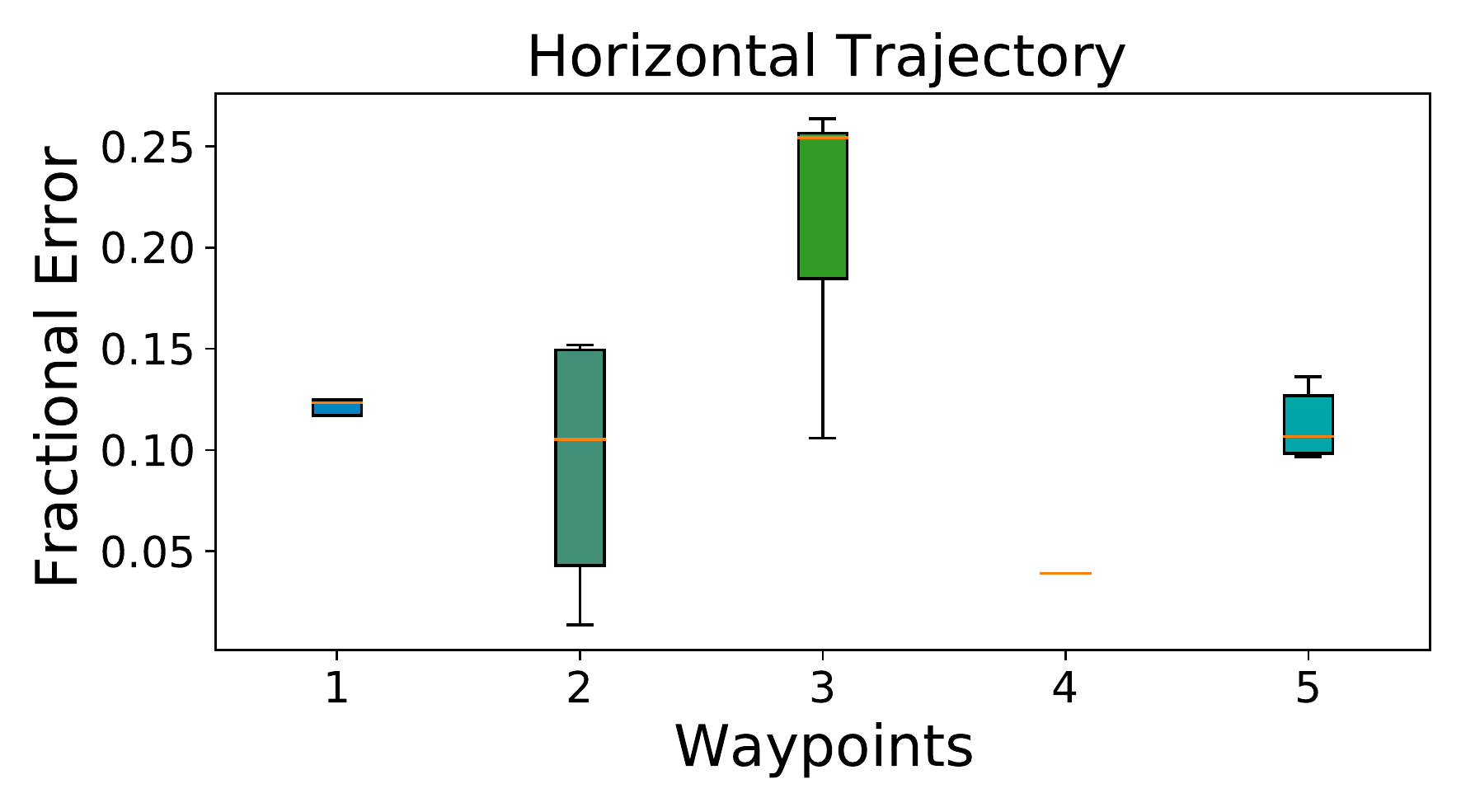}
        (B)
    \end{minipage}\hfill%
    \begin{minipage}[b]{0.5\textwidth}
        \centering
        \includegraphics[width=1.0\linewidth]{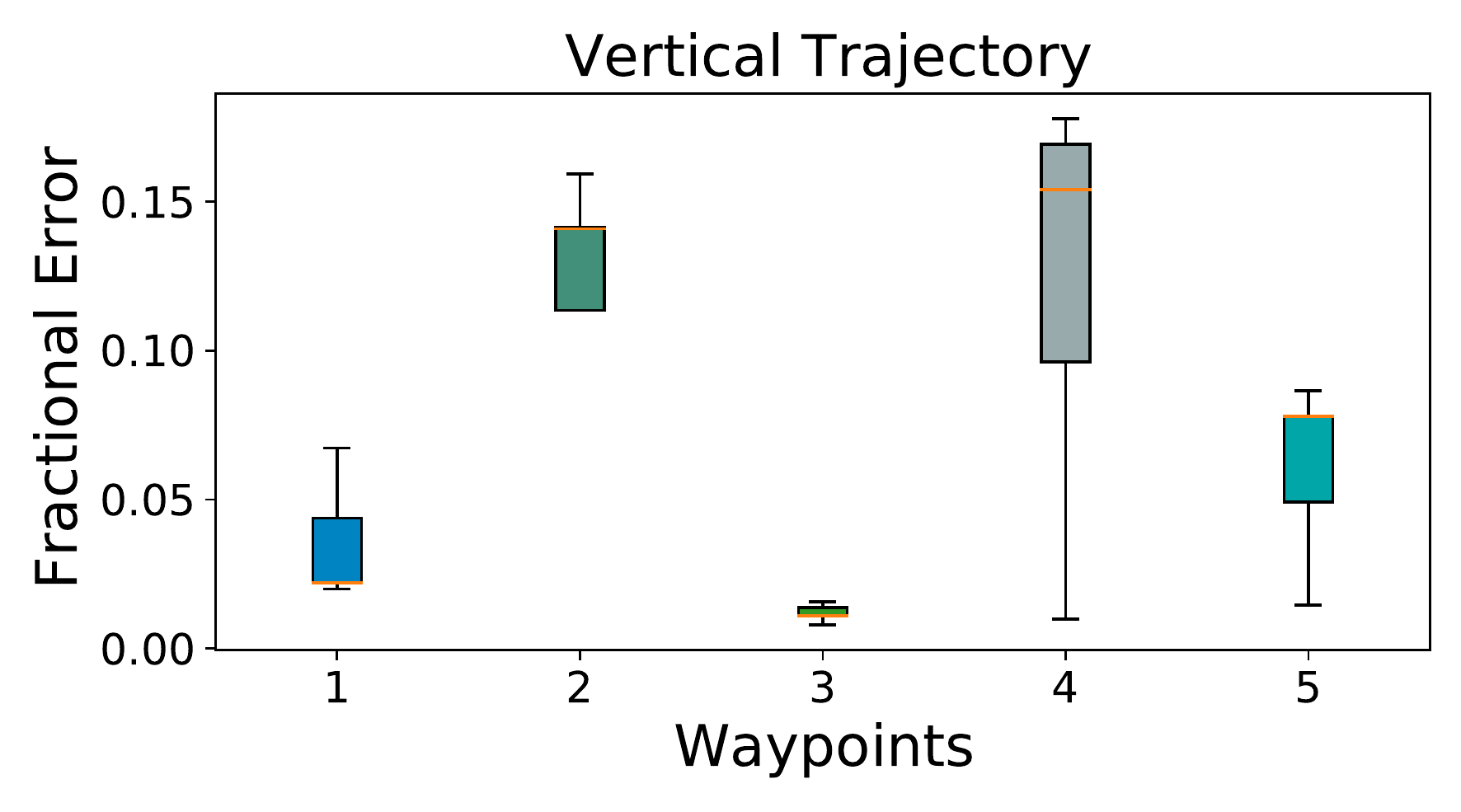}
        (B)
    \end{minipage}\\
    \begin{minipage}[t]{0.5\textwidth}
        \centering
        \includegraphics[width=1.0\linewidth]{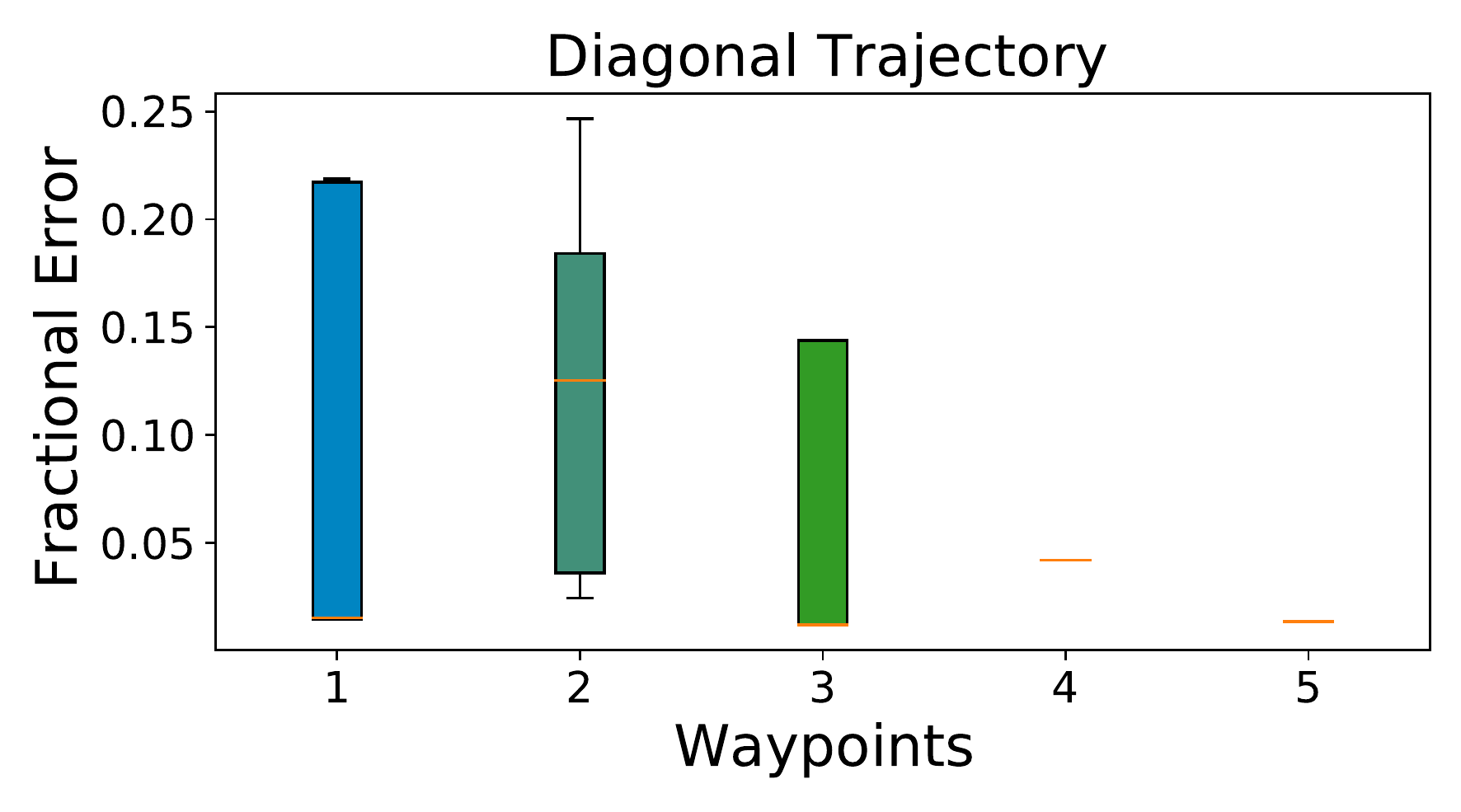}
        (C)
    \end{minipage}\hfill%
    \begin{minipage}[t]{0.5\textwidth}
        \centering
        \includegraphics[width=1.0\linewidth]{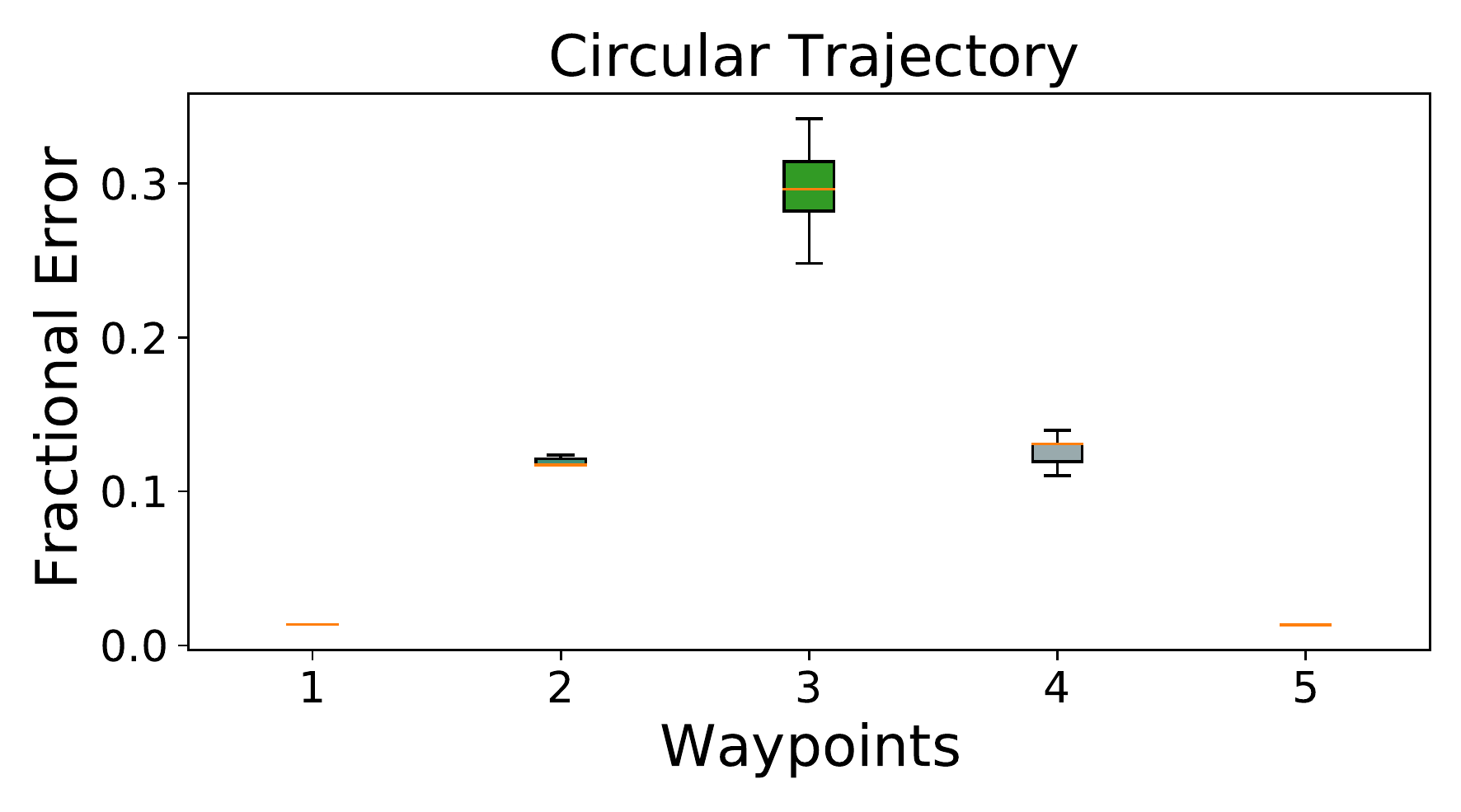}
        (D)
    \end{minipage}
    \caption{Touch Tracking Error Plots. The middle bar represents the median error, the width of each bar is the Interquartile Range, and the fence widths are $1.5\times\textrm{IQR}$. (A) Average Deviation for Vertical Trajectory, (B) Average Deviation for Horizontal Trajectory, (C) Average Deviation for Diagonal Trajectory, (D) Average Deviation for Circular Trajectory}
    \label{fig:touch_tracking_errors}
\end{figure*}

We collected data from the \btsp at a rate of 100 Hz by making contact at different sensor locations. Three different probes with varying indenter diameters (1, 2 and 5 mm respectively) were used to gather this dataset. The taxel values are time-synchronised with an RGB camera feed which provides us with visual ground truth of contact location at every instance. This data was then used to generate events according to the method described in Sec.~\ref{subsec:tactile_gen}.

We evaluate our method of tracking contact by comparing it qualitatively with the ground truth trajectories of the probes obtained from the RGB images. We hand-label several marker locations (shown in Fig.~\ref{fig:touch_tracking_markers} on the physical sensor and align them in image coordinate space to the 2D projected locations of the taxels. We used 8 different trajectories, as shown in Fig.~\ref{fig:touch_tracking}.

We move the indenters on various trajectories along the surface of the \btsp, as shown in Fig.~\ref{fig:touch_tracking}, from top to bottom, bottom to top (Fig.~\ref{fig:touch_tracking}(A)), left to right, right to left ((Fig.~\ref{fig:touch_tracking}(B)), diagonally top to bottom, diagonally bottom to top (Fig.~\ref{fig:touch_tracking}(C)), circular clockwise, and circular counter-clockwise (Fig.~\ref{fig:touch_tracking}(D)).

Fig.~\ref{fig:touch_tracking_plots} shows the outputs from two sample trajectories  --  diagonal motion from bottom left to top right and counter-clockwise circular motion. Fig.~\ref{fig:touch_tracking_plots}(A) and Fig.~\ref{fig:touch_tracking_plots}(C) show the trajectories overlaid on the  contour surfaces generated from the event aggregates stacked along the time axis. In both outputs, we can clearly see the event aggregates in red representing the current region of touch. Tracking these across time, we can generate a trajectory of touch across the skin surface.\\
For comparison, in Figs.~\ref{fig:touch_tracking_plots}(B) and \ref{fig:touch_tracking_plots}(D) we compare the outputs obtained from the filtered, but otherwise unprocessed raw data from the \btsp. It is clear that the outputs from our approach, shown in in Fig.~\ref{fig:touch_tracking_plots}, produces smoother trajectories with reduced noise.
 Table~\ref{table:contact-location-errors} provides  a comparison of the average pixel-wise error in tracking the known waypoints (Fig.~\ref{fig:touch_tracking_markers}), computed with three different methods: First, we take the 24 taxel values corresponding to the timestamp at which the indenter is on each of the known waypoints 1 through 5, and train a fully connected neural network for regression on predicted locations. The average pixel-wise errors are reported under the \textit{Raw MLP} heading.
Similarly, we obtain the average pixel-wise errors for each of the five waypoints, using the contours from raw data and from event data. These are reported under the \textit{Raw Contours} and \textit{Event Contours} headers respectively in Table~\ref{table:contact-location-errors}.

\begin{table}[]
\centering
\begin{tabular}{c c c c c c c}
\toprule
\multicolumn{2}{c}{} & \textbf{1} & \textbf{2} & \textbf{3} & \textbf{4} & \textbf{5}\\
\midrule
\multirow{3}{*}{\textbf{Circle}}
& \textit{Raw MLP} & 0.76 & 0.47 & 0.78 & 0.45 & 0.75 \\
& \textit{Raw Contours} & 0.07 & \textbf{0.06} & 0.35 & 0.15 & 0.31 \\
& \textit{Event Contours} & \textbf{0.04} & 0.11 & \textbf{0.29} & \textbf{0.12} & \textbf{0.02}\\
\midrule
\multirow{3}{*}{\textbf{Diagonal}}
& \textit{Raw MLP} & 0.86 & 0.30 & \textbf{0.01} & 0.28 & 0.75 \\
& \textit{Raw Contours} & \textbf{0.02} & 0.24 & 0.17 & 0.05 & 0.02 \\
& \textit{Event Contours} & 0.10 & \textbf{0.11} & 0.05 & \textbf{0.05} & \textbf{0.01}\\
\midrule
\multirow{3}{*}{\textbf{Horizontal}}
& \textit{Raw MLP} & 0.47 & 0.33 & \textbf{0.01} & 0.30 & 0.45 \\
& \textit{Raw Contours} & \textbf{0.04} & 0.10 & 0.32 & 0.40 & 0.50 \\
& \textit{Event Contours} & 0.13 & \textbf{0.09} & 0.21 & \textbf{0.04} & \textbf{0.01}\\
\midrule
\multirow{3}{*}{\textbf{Vertical}}
& \textit{Raw MLP} & 0.56 & 0.28 & \textbf{0.02} & 0.22 & 0.40 \\
& \textit{Raw Contours} & 0.59 & 0.32 & 0.17 & 0.14 & 0.10 \\
& \textit{Event Contours} & \textbf{0.05} & \textbf{0.11} & \textbf{0.02} & \textbf{0.12} & \textbf{0.07}\\
\bottomrule
\end{tabular}
\caption{Mean errors in the ratio of computed contact location and \btsp width at each way-point over different trajectories}
\label{table:contact-location-errors}
\end{table}
\graphicspath{{\subfix{../images/}}}

\subsection{Magnitude of Force}\label{subsec:magnitude_force}
We applied varying forces on the \btsp skin using a 2mm indenter  to demonstrate the ability of our event contours to measure the correlation between the magnitude of contact force and the area of the maximal contour. The ground truth forces were measured with the help of a calibrated and accurate force sensor.

In order to obtain the relationship between contour areas and applied force, we trained a fully connected neural network with 7 hidden layers, with layer widths of 8, 16, 32, 64, 32, 16 and 8 respectively using  L2 loss. This network was used to compute a regression curve mapping the forces to the contour areas. We applied a logistic activation function, and used an inversely scaled learning rate. The network was trained for 5000 epochs, on 200 data points.\\
As a point of comparison, we applied two other regression methods, a stochastic gradient descent regression with the ElasticNet loss function, and another L2 regression with Huber loss.\\
We used the mean absolute percentage error, defined as 
\begin{equation}
    \mathrm{MAP}(y,\hat{y}) = \mathbb{E} \frac{|y_i - \hat{y}|}{\mathrm{max(\epsilon, |y_i|)}}
\end{equation}
where $\mathbb{E}$ is the expectation operator.

\begin{figure}[h!]
\centering
\includegraphics[width=1.0\linewidth]{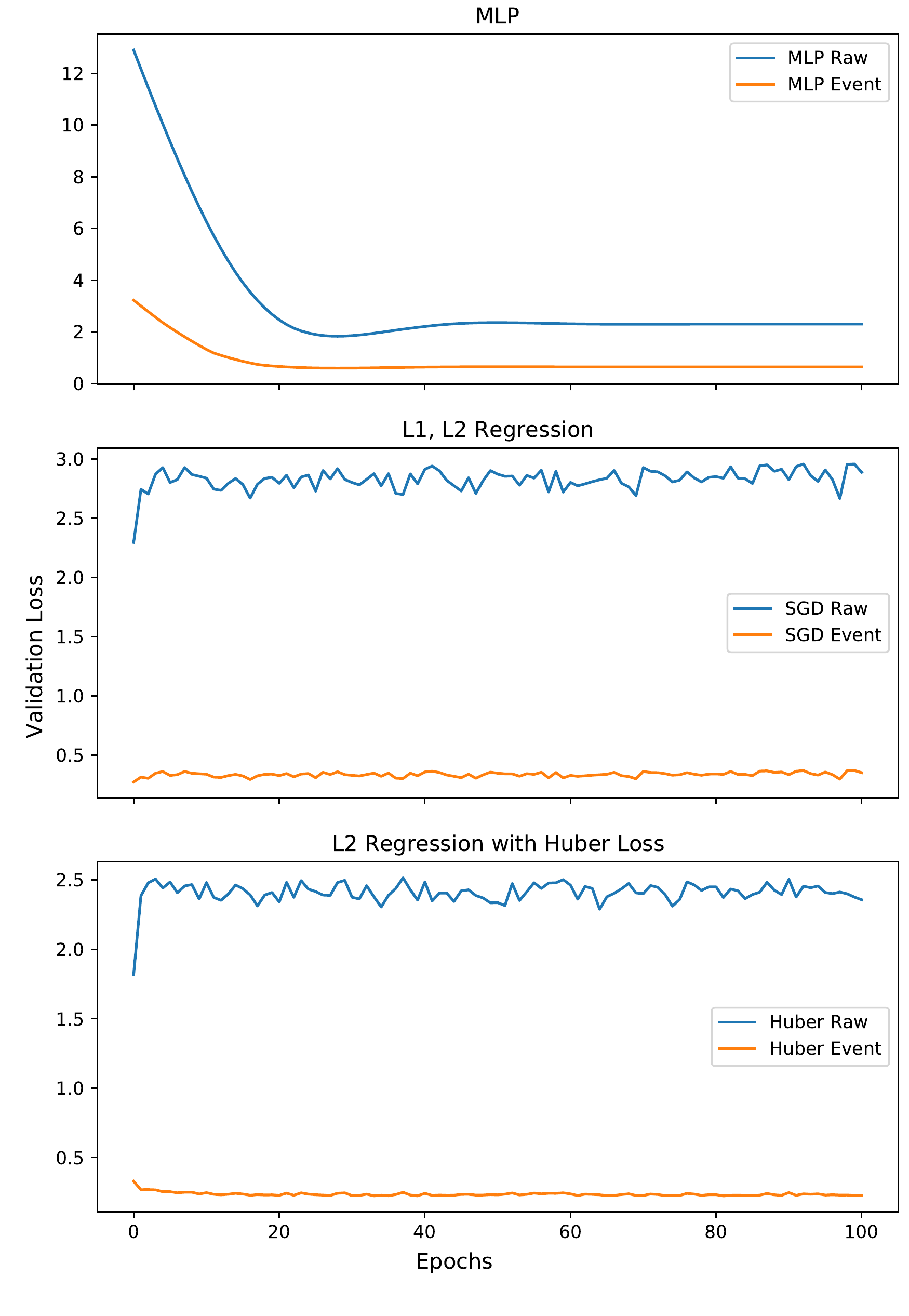}
\caption{Comparing various force-area regression methods for raw vs. event data.}
\label{fig:force-area-plot}
\end{figure}

Each of these  methods were trained for 1000 epochs over 200 data points, but for brevity in Fig.~\ref{fig:force-area-plot} we only display 100 epochs. The figure also shows the results from the same regression techniques applied to the contours generated from raw data. It is evident from the plots, across all learning algorithms, that the event based data in comparison to the raw data, shows a better validation loss curve during training, and has an overall lower loss score at testing.

\begin{figure*}[h!]
    \centering
     \begin{minipage}[t]{0.24\textwidth}
        \centering
        \includegraphics[width=1.0\linewidth]{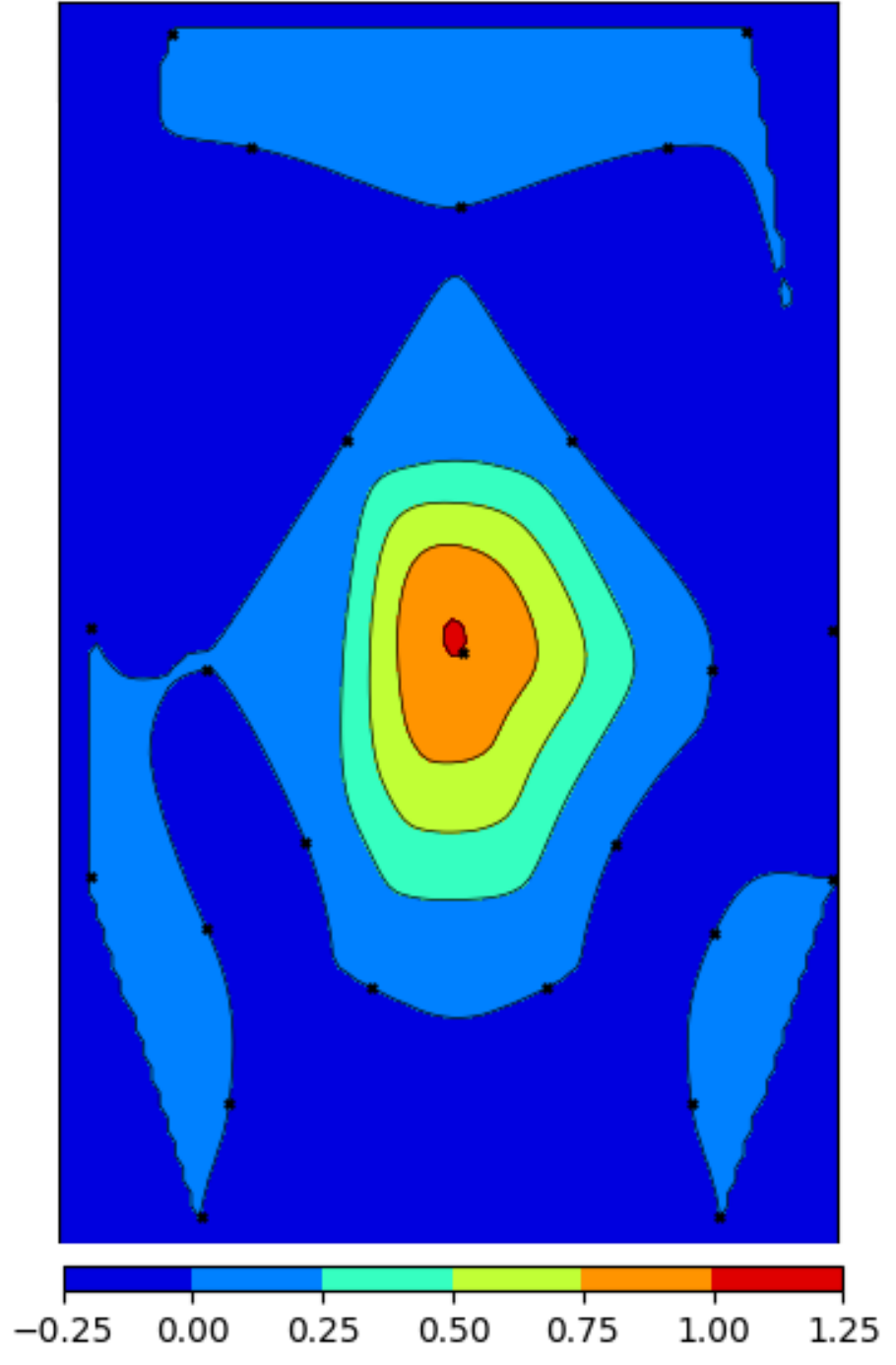}
        (A)
    \end{minipage}\hfill%
     \begin{minipage}[t]{0.23\textwidth}
        \centering
        \includegraphics[width=1.0\linewidth]{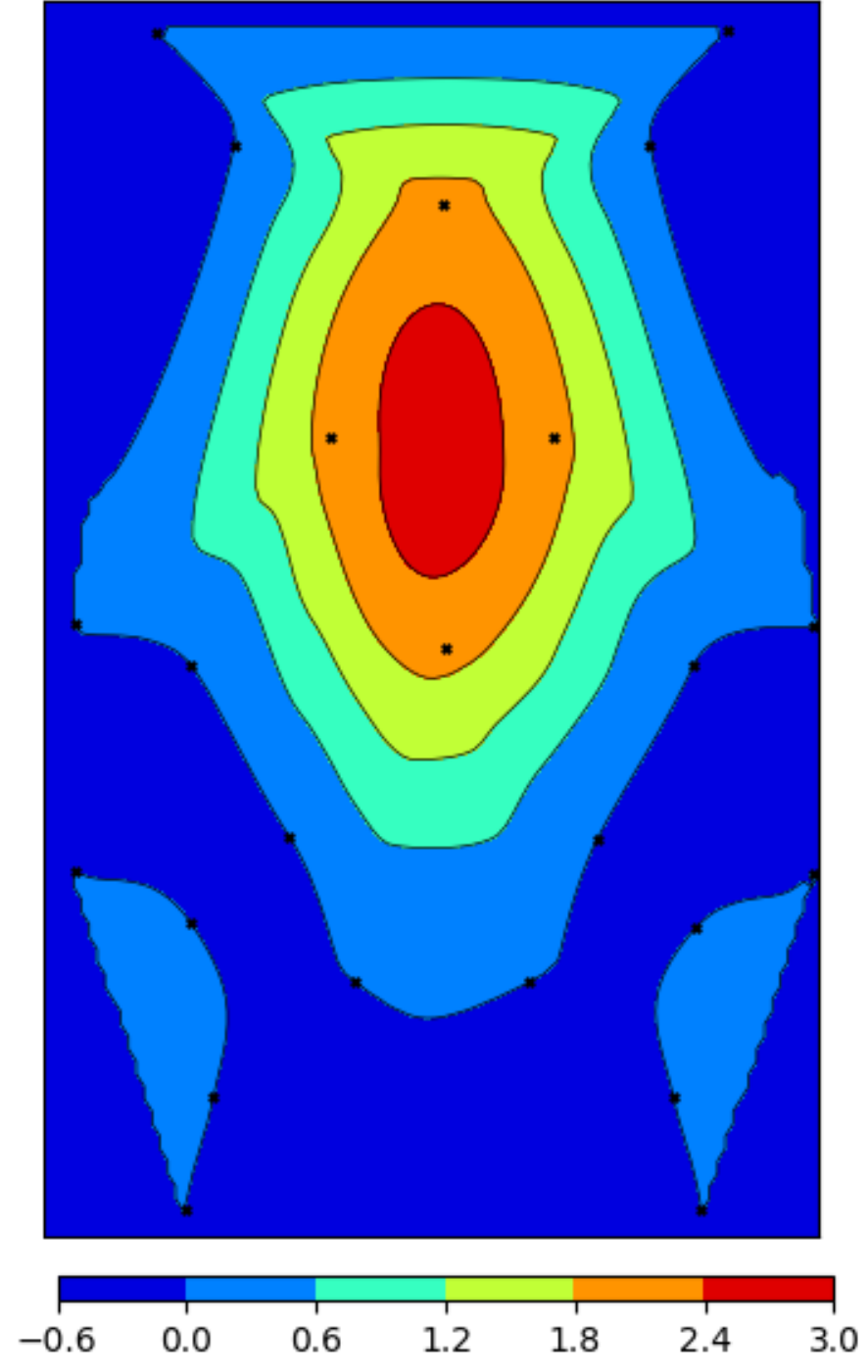}
        (B)
    \end{minipage}\hfill%
     \begin{minipage}[t]{0.22\textwidth}
        \centering
        \includegraphics[width=1.0\linewidth]{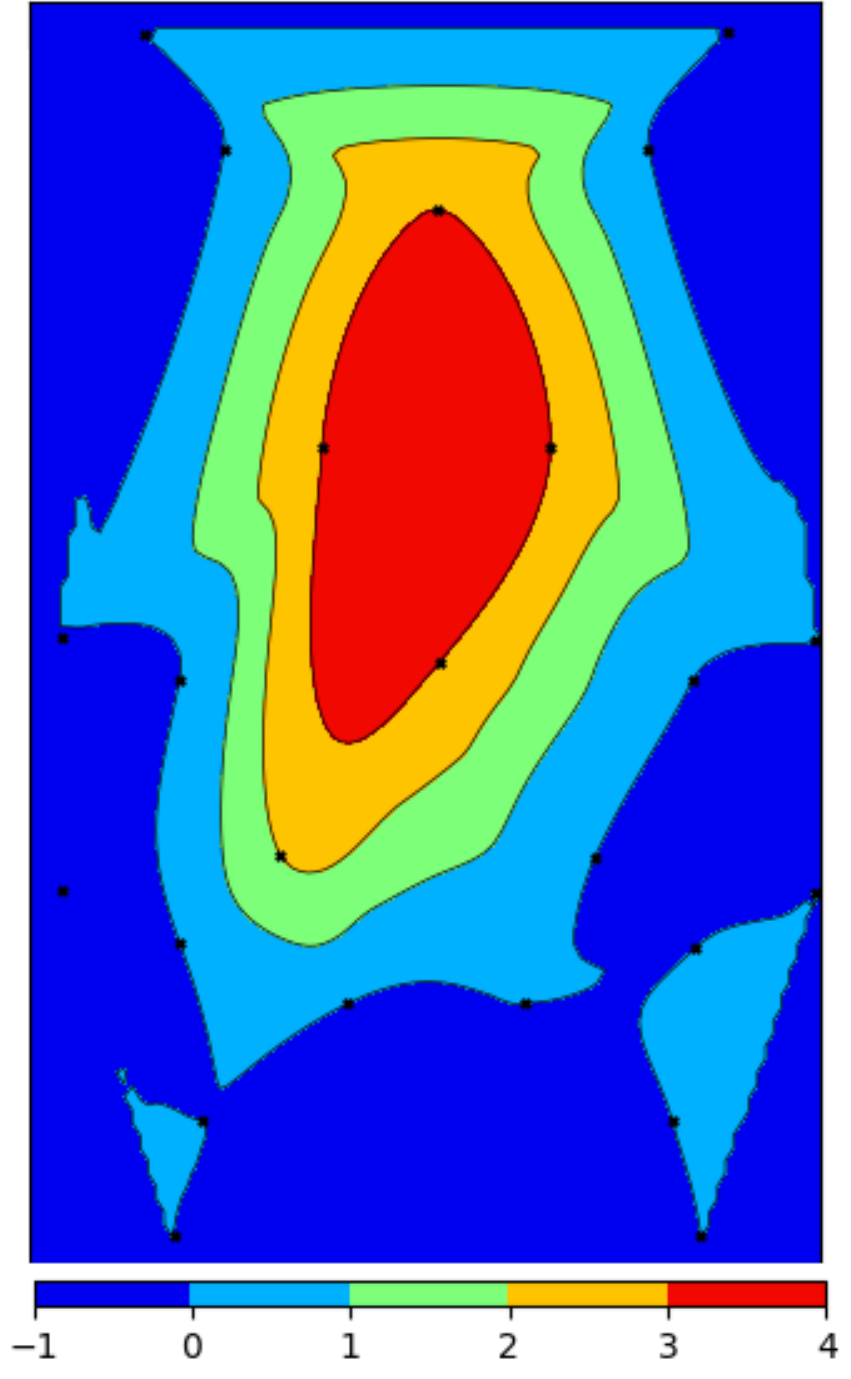}
        (C)
    \end{minipage}
    \label{fig:force-area}
    \caption{Contour region areas correlated with applied force. (A) 3N applied force, (B) 6N applied force, (C) 12N applied force}
\end{figure*}
\graphicspath{{\subfix{../images/}}}

\subsection{Slippage Detection and Classification}\label{subsec:slippage_detection}
\begin{figure*}[h!]
    \centering
    \begin{minipage}[t]{0.33\textwidth}
        \centering
        \includegraphics[width=1.0\linewidth]{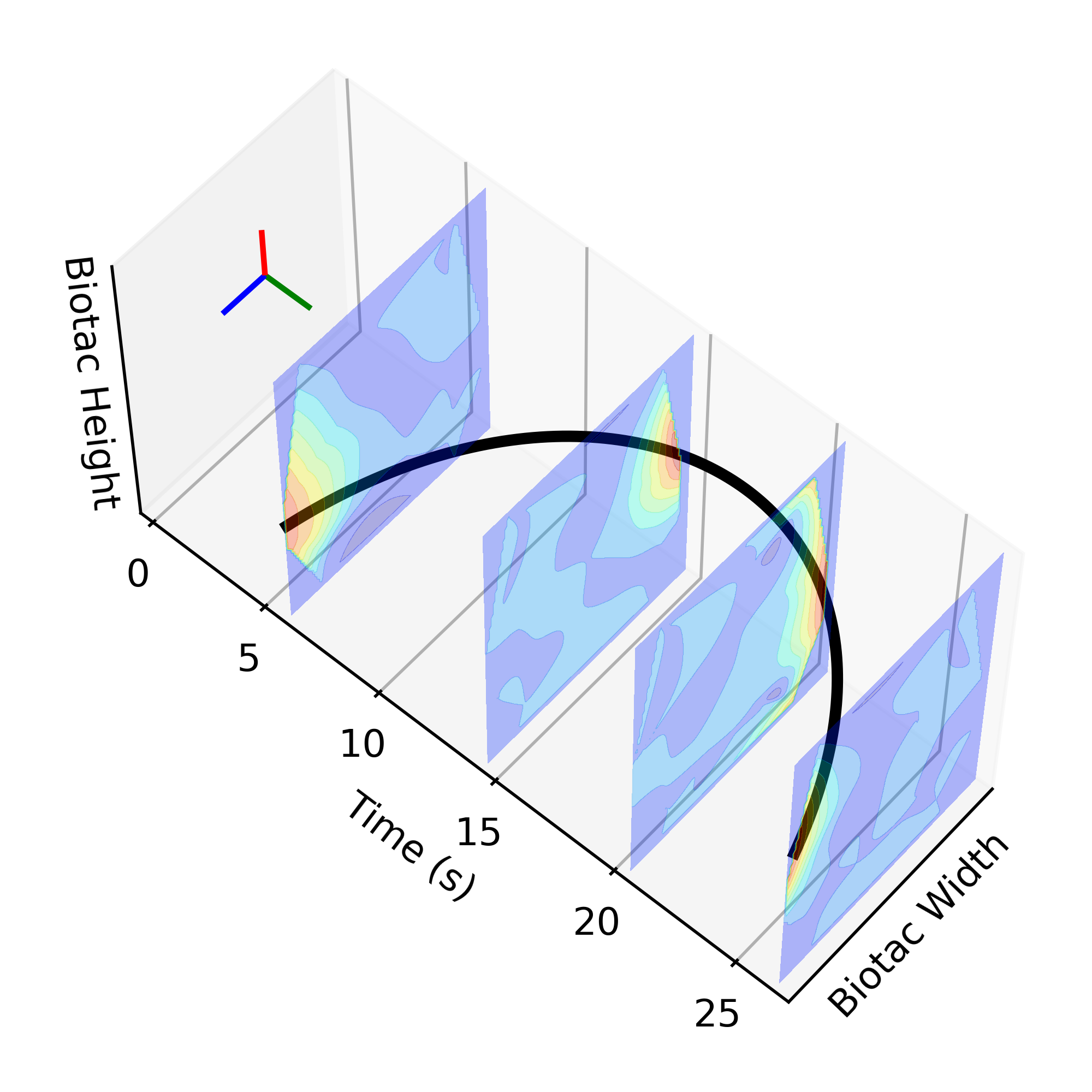}
        (A)
    \end{minipage}\hfill%
    \begin{minipage}[t]{0.33\textwidth}
        \centering
        \includegraphics[width=1.0\linewidth]{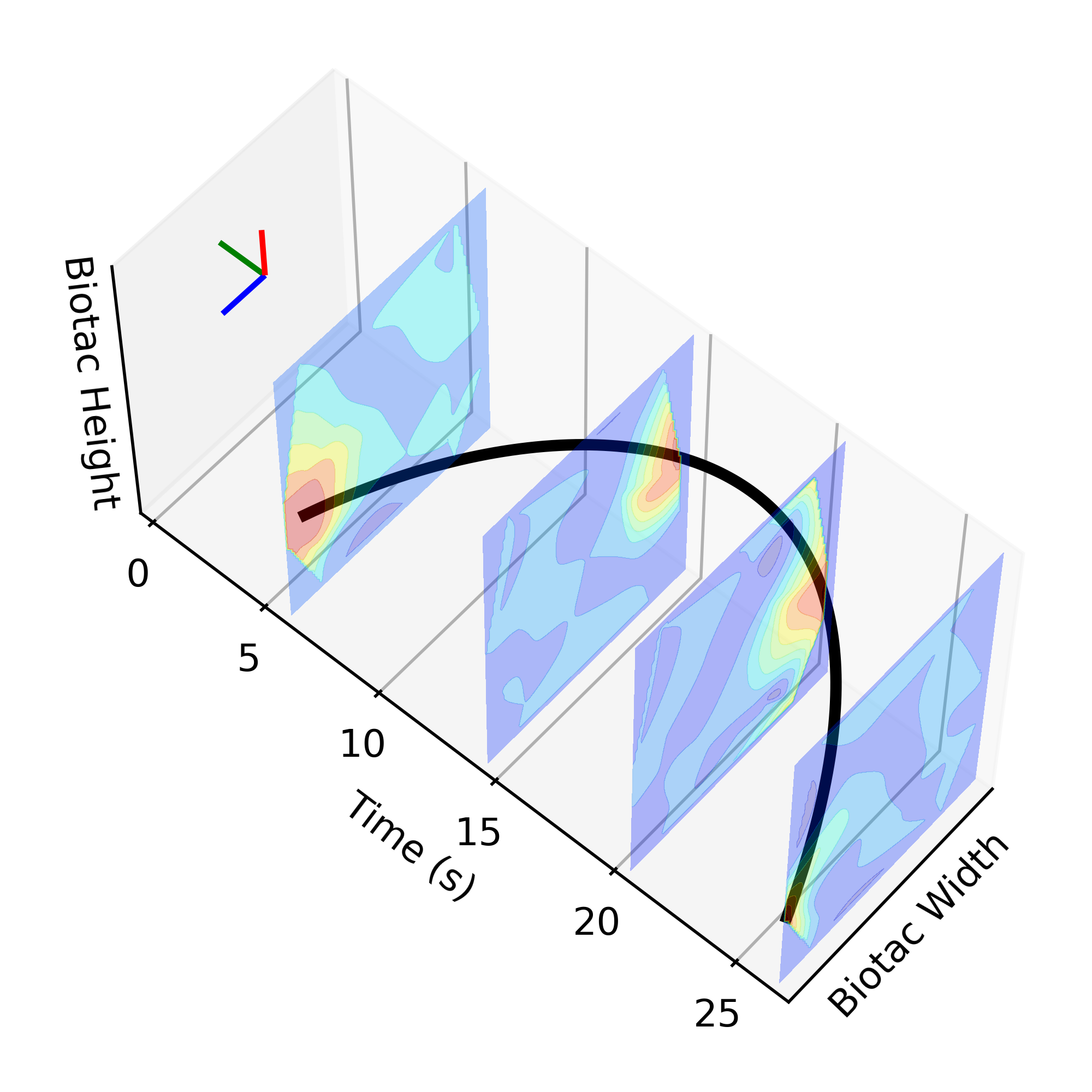}
        (B)
    \end{minipage}\hfill%
    \begin{minipage}[t]{0.25\textwidth}
        \centering
        \includegraphics[width=1.0\linewidth]{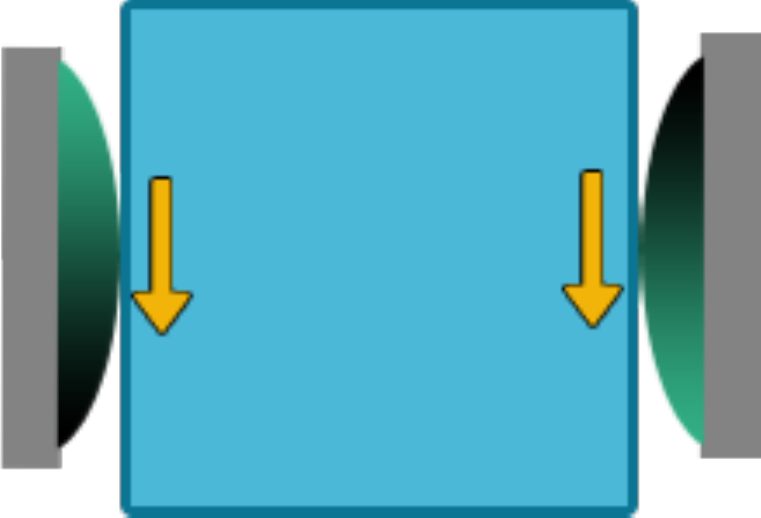}
        (C)
    \end{minipage}\\
    \begin{minipage}[t]{0.33\textwidth}
        \centering
        \includegraphics[width=1.0\linewidth]{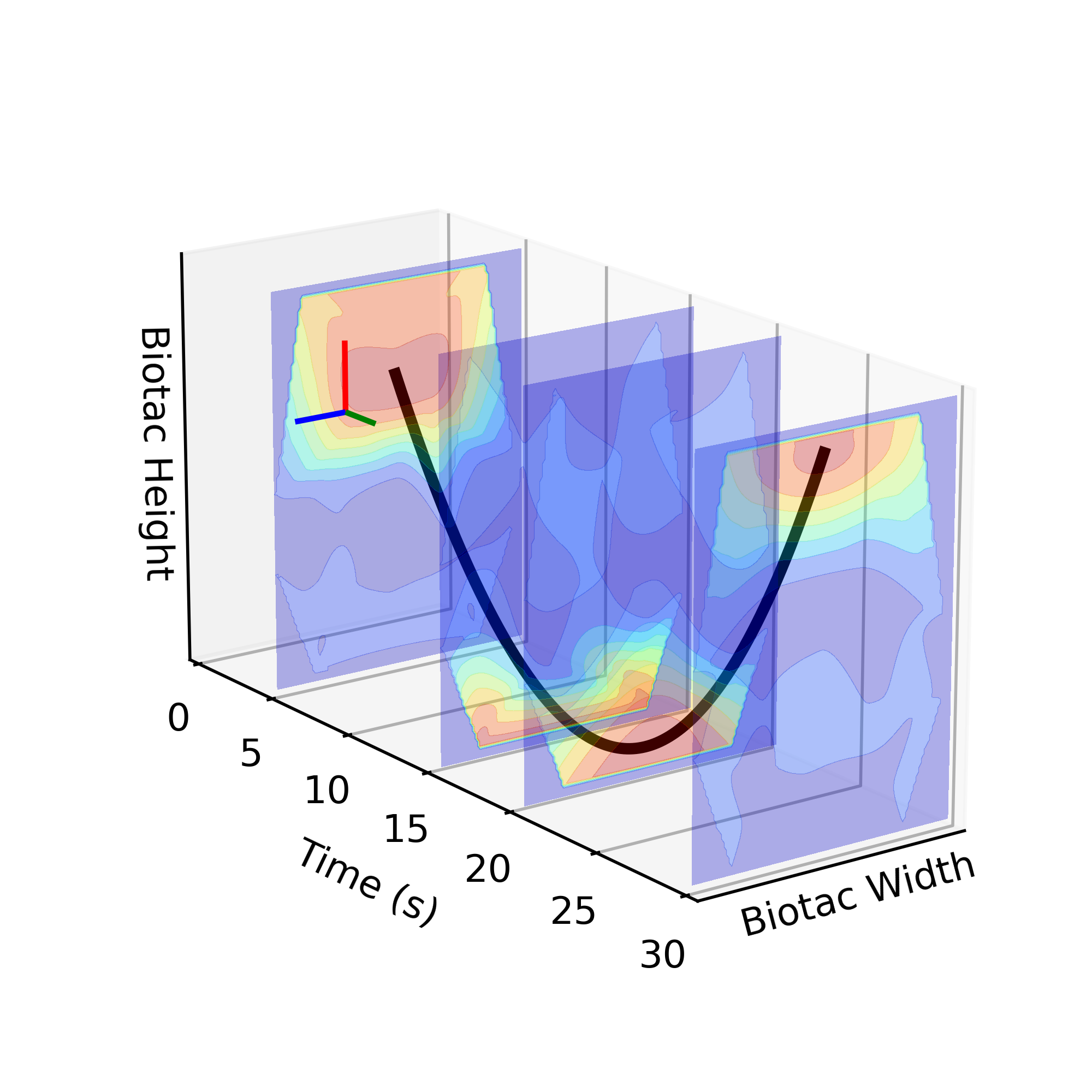}
        (D)
    \end{minipage}\hfill%
    \begin{minipage}[t]{0.33\textwidth}
        \centering
        \includegraphics[width=1.0\linewidth]{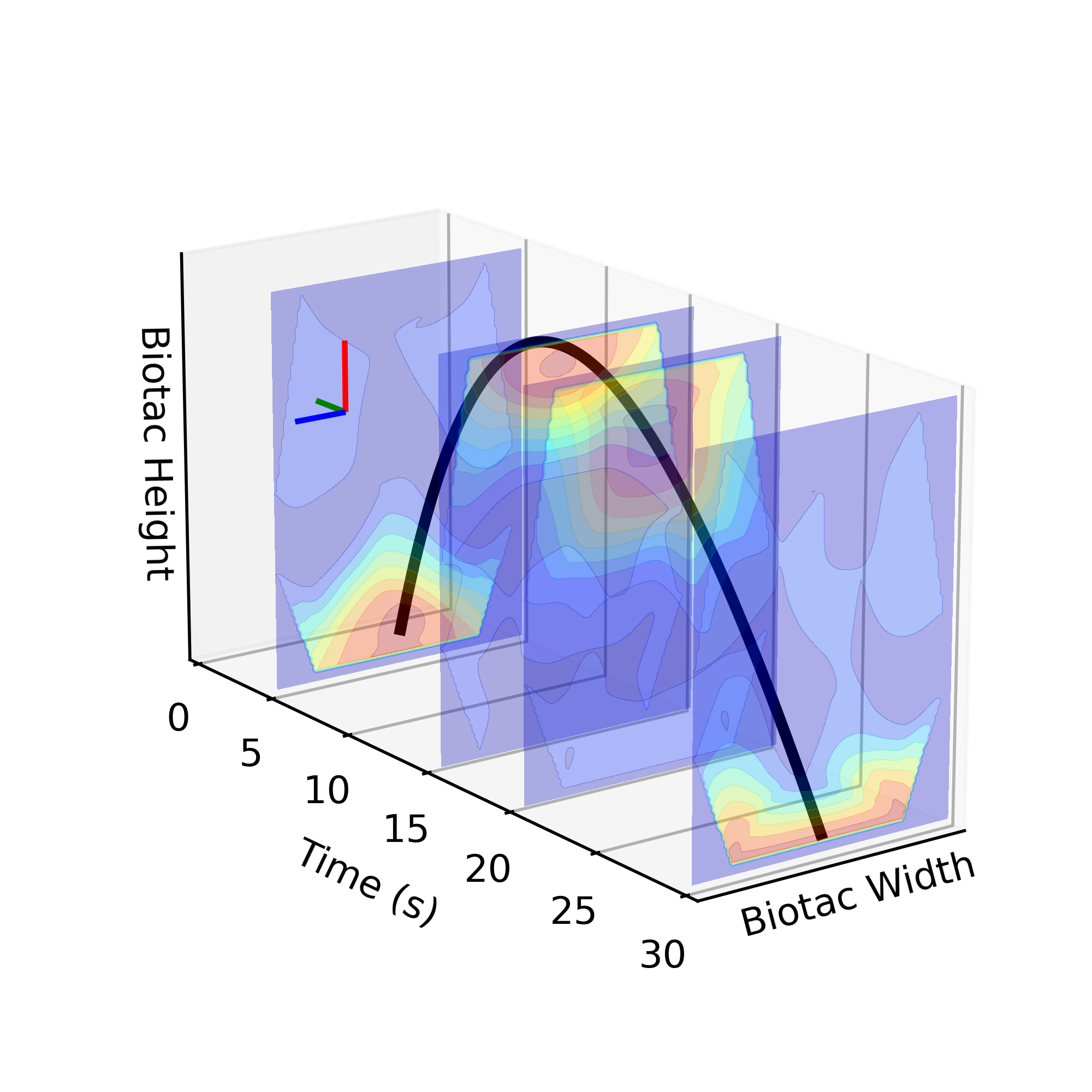}
        (E)
    \end{minipage}\hfill%
    \begin{minipage}[t]{0.25\textwidth}
        \centering
        \includegraphics[width=1.0\linewidth]{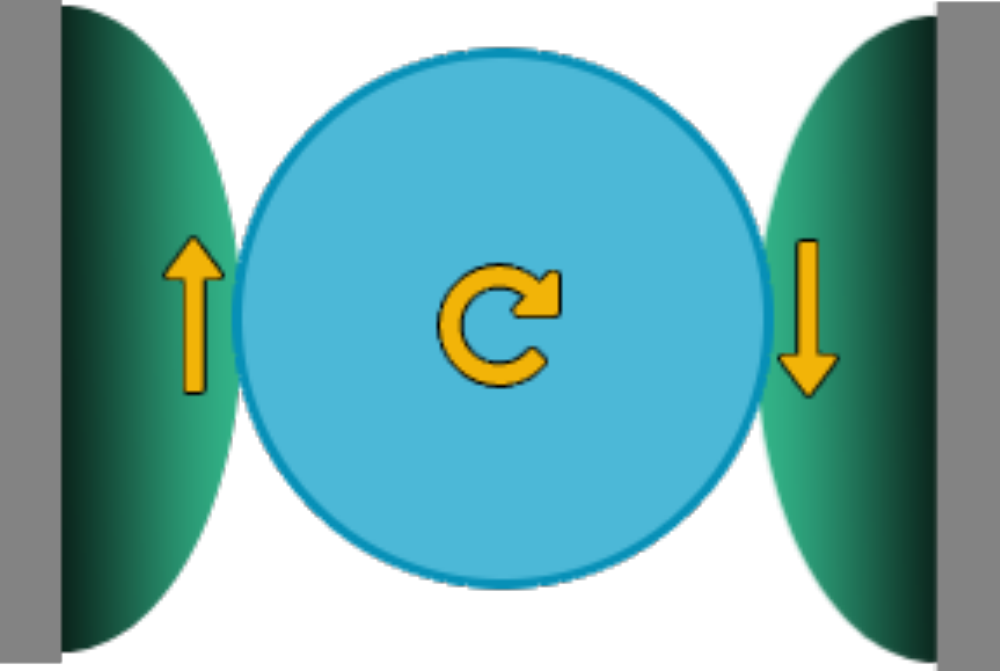}
        (F)
    \end{minipage}\\
    \caption{Examples of trajectories during longitudinal and rotational slippage. (A), (B) First Finger and Thumb trajectories for longitudinal slippage, (C) Directional Diagram for longitudinal slippage, (D), (E) First Finger and Thumb trajectories for rotational slippage, (F) Directional Diagram for rotational slippage}
    \label{fig:slip_diagrams}
\end{figure*}

 \begin{figure}[h!]
    \centering
    \includegraphics[width=1.0\linewidth]{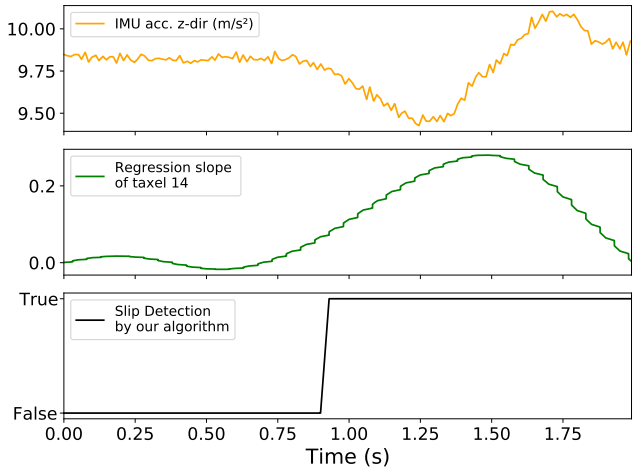}
    \caption{Slip Detection Comparison Plots. From top to bottom, we have a low-pass filtered acceleration on the z-axis, the regression slope on the raw data, and the binary slip detection results from event contours.}
    \label{fig:slip_imu}
\end{figure}

\begin{figure}[h!]
    \centering
    \begin{minipage}[t]{0.24\textwidth}
        \centering
        \includegraphics[width=1.0\linewidth]{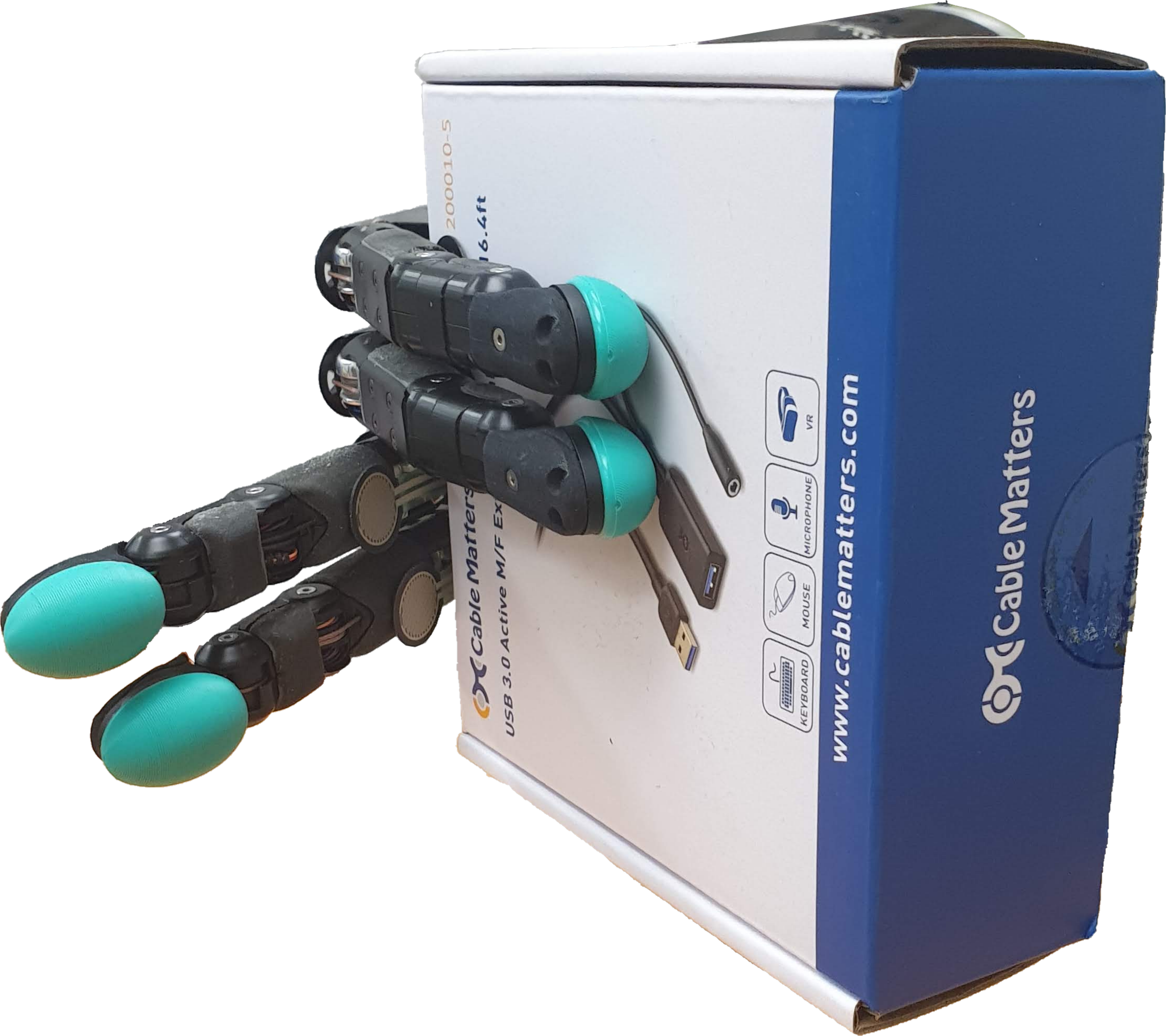}
        (A)
    \end{minipage}\hfill%
    \begin{minipage}[t]{0.24\textwidth}
        \centering
        \includegraphics[width=1.0\linewidth]{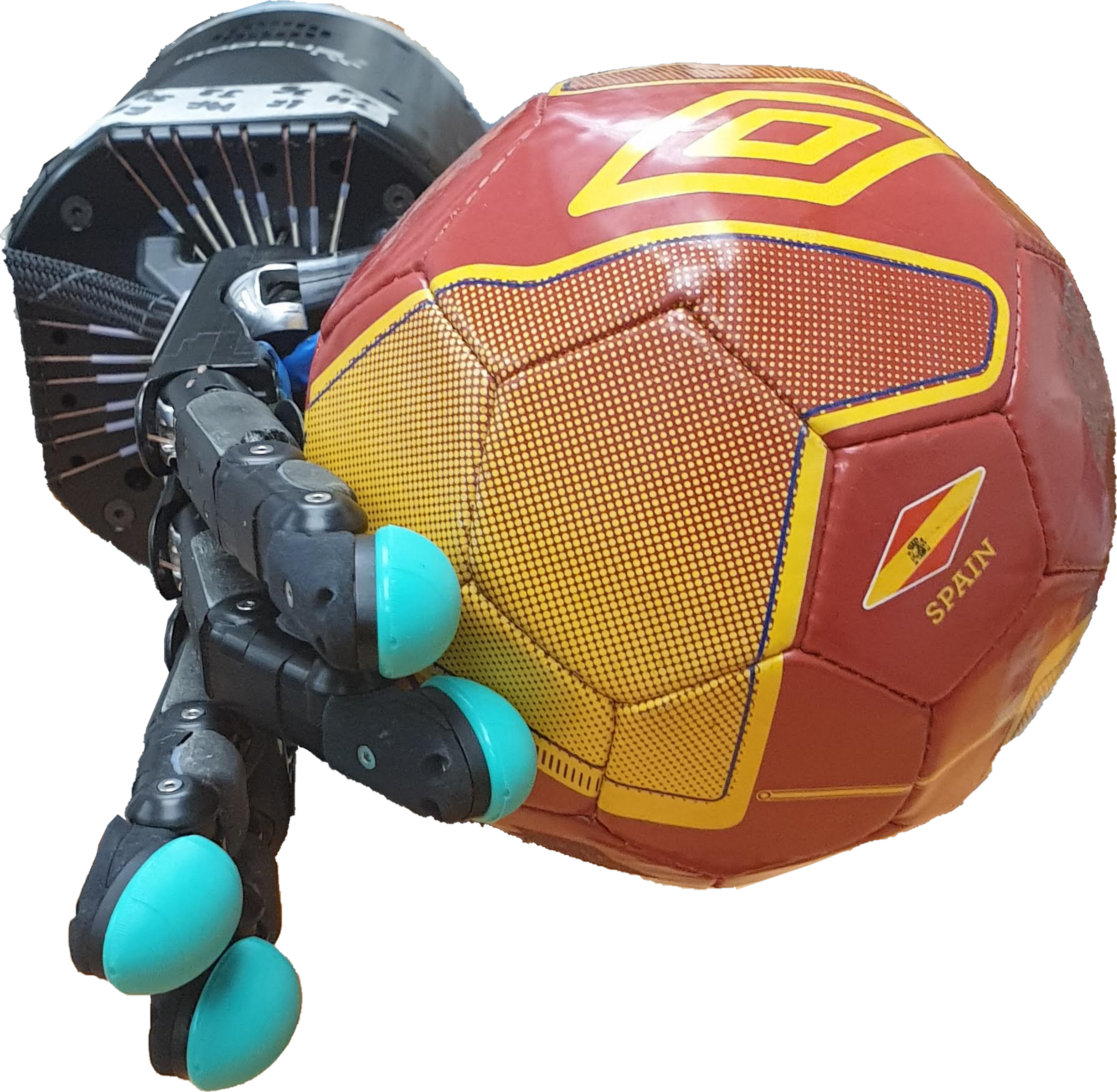}
        (B)
    \end{minipage}\hfill\\
    \begin{minipage}[t]{0.24\textwidth}
        \centering
        \includegraphics[width=1.0\linewidth]{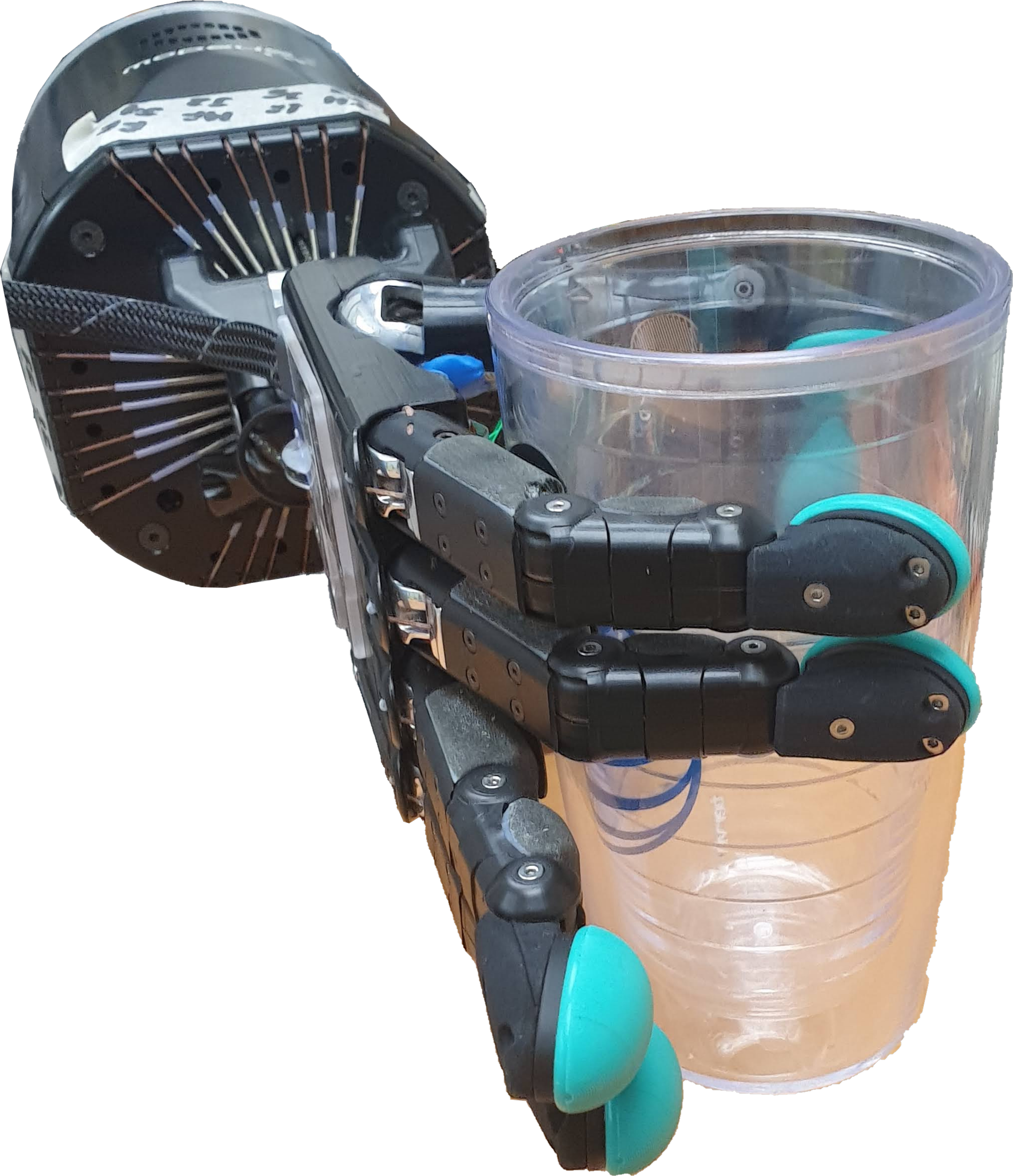}
        (C)
    \end{minipage}\hfill%
    \begin{minipage}[t]{0.20\textwidth}
        \centering
        \includegraphics[width=1.0\linewidth]{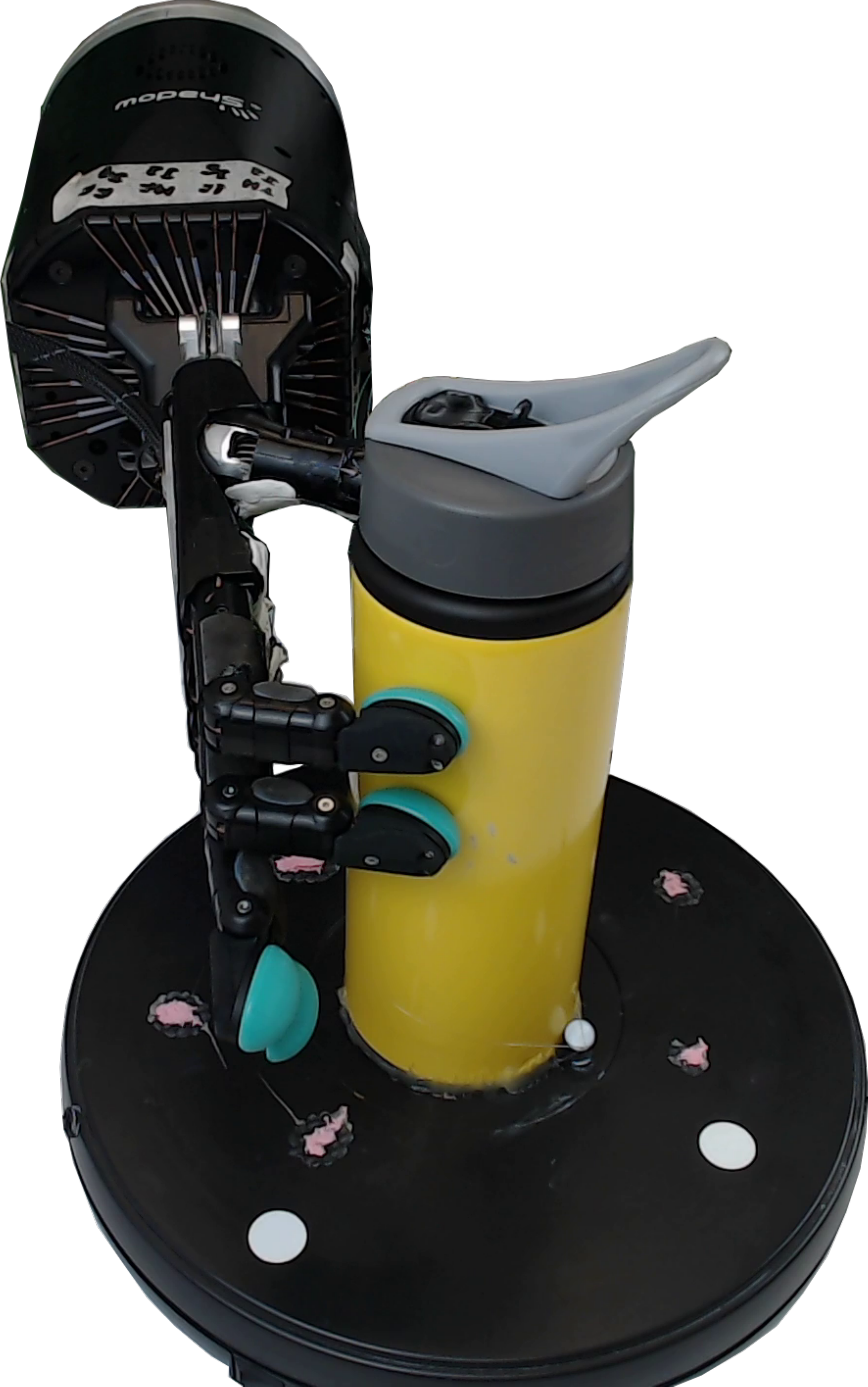}
        (D)
    \end{minipage}
    \caption{Objects Used for Longitudinal and Rotational Slip Detection. (A) Box shape, (B) Spherical shape, (C) Cylinder shape, (D) Tumbler on constant-speed turntable}
    \label{fig:slip_objects}
\end{figure}

There are many %distinct ways
different ways 
 slippage detection has been achieved using the \btsp (\cite{schaal2015force}, \cite{veiga2020tactilegripstabilization}, \cite{calandra2018more}, \cite{naeini2019novel}), with most methods specifically designed for the task.
Here we show that our generic method of spatio-temporal contours can also be used for slippage detection and classification, demonstrating that our approach is very adaptive.
%This makes our approach more adaptive than other purpose-built algorithms for slippage detection from tactile data.

By tracking the contours spatio-temporally, we are able to detect both the time at which slippage occurs, as well as its directionality. In case of longitudinal slip, i.e. in which the object moves linearly between the fingers, we can measure the direction as ``up'' or ``down''. In case of rotational slip, i.e. in which the object rotates between the fingers, we can measure clockwise vs. counter-clockwise rotation.

We do this by comparing the event contours from fingers on opposing sides of the object, while the object is fully grasped by the Shadow Hand, as shown in Figs.~\ref{fig:slip_objects}.
By tracking and comparing the trajectories generated by the contours on the first finger and the thumb, we can deduce both the time at which slippage occurs, as well as the direction.
In case of longitudinal slippage, as in Fig.~\ref{fig:slip_diagrams}(C), based on the orientation of the \btsp sensors with respect to the object, both the contour trajectories have the same direction of motion.
In case of rotational slippage, as in Fig.~\ref{fig:slip_diagrams}(F), because of opposing shear forces experienced on the thumb versus the first finger, the contour trajectories  have opposing directions of motion.

For the longitudinal slippage scenario, the object is allowed to slide down and is then gradually  pulled back up, while maintaining a stable grasp. This can be seen by the contours moving from left to right spatially across the sensor's surface, and then from right back to the left. As is evident from the trajectories, because of shear forces being in the same direction for both sensors, the direction of the respective trajectories are also the same.
For the rotational slippage scenario, the object is affixed to a constant-speed turntable and allowed to rotate slowly between the opposing fingers. In this motion, due to the resultant opposing shear forces, the contour trajectories for the index finger and the thumb have clearly opposite directions.

The event contour outputs of these experiments, longitudinal and rotational slippage, are shown in Figs.~\ref{fig:slip_diagrams}(A),(B) and \ref{fig:slip_diagrams}(C),(D) respectively. In both cases the contours on the \btsp sensor are tracked over time, separately for the index finger and the thumb, which as per the diagrams (Figs.~\ref{fig:slip_diagrams}(A),(B) and \ref{fig:slip_diagrams}(C),(D)) have different  orientations.\\
We obtain ground truth for our experiments using an 6-axis IMU mounted on each object, and use time synchronized outputs from the IMU to compute the time of slip. We compare our event-based approach to a regression slope computed on the raw data, and the results of one such experiment is shown in Fig.~\ref{fig:slip_imu}.
\graphicspath{{\subfix{../images/}}}

\subsection{Tracking Edges using Contact Location}
\label{subsec:tracking_edges_results}

\begin{figure}[ht!]
    \centering
    \begin{minipage}[t]{0.24\textwidth}
        \centering
        \includegraphics[width=1.0\linewidth]{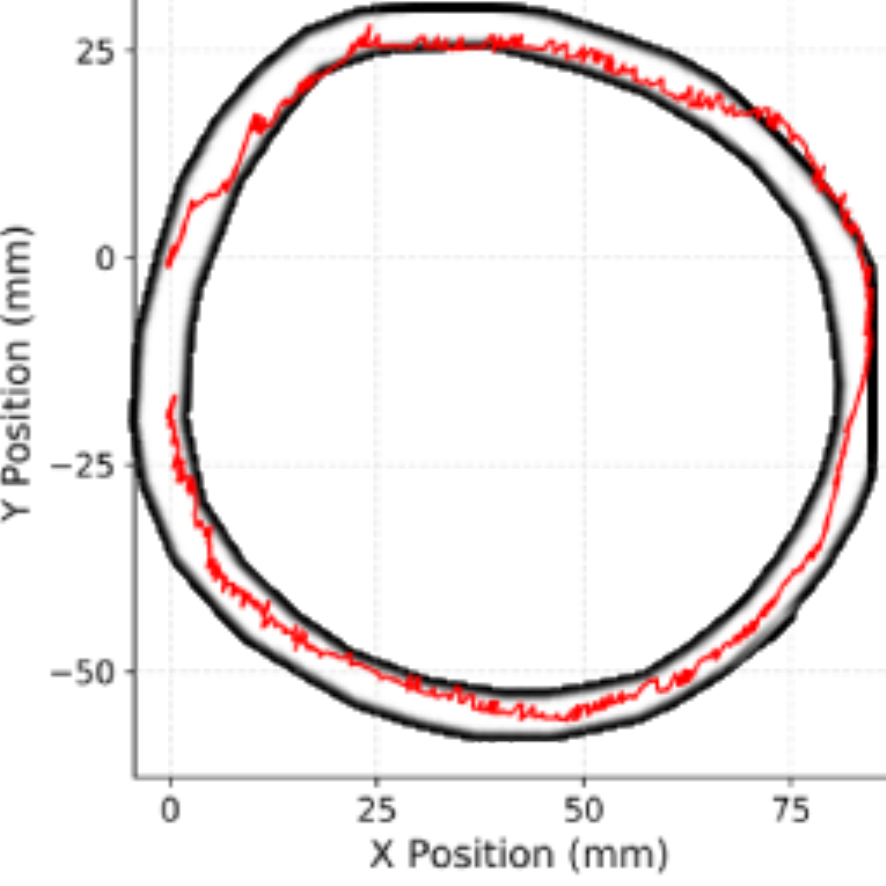}
        (A)
    \end{minipage}\hfill%
    \begin{minipage}[t]{0.24\textwidth}
        \centering
        \includegraphics[width=1.0\linewidth]{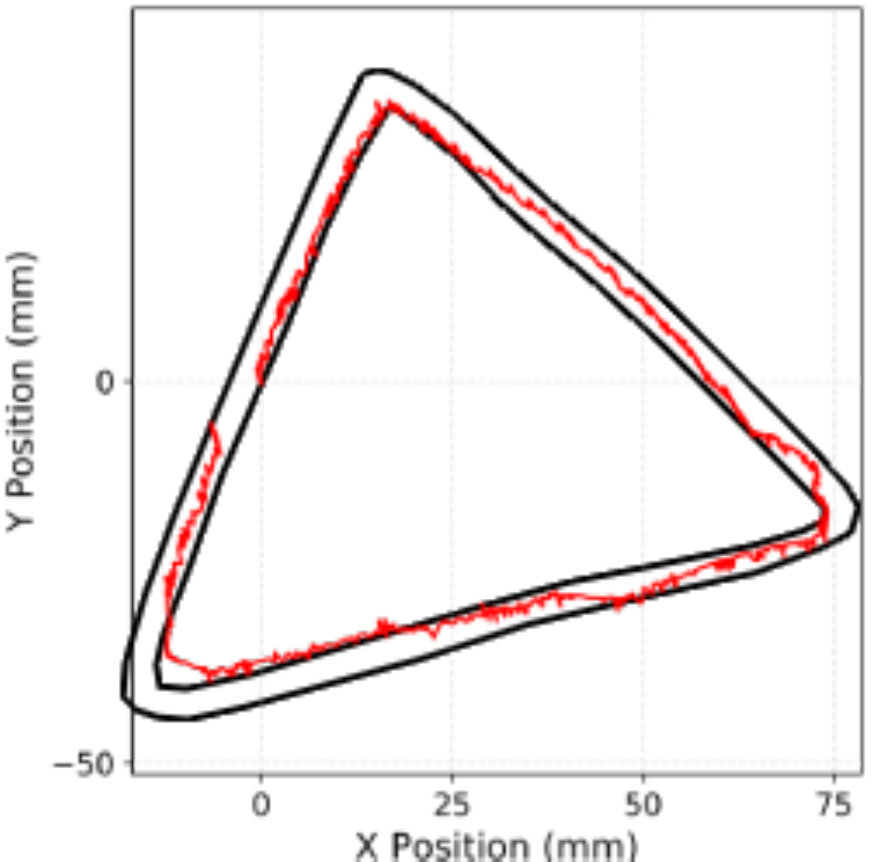}
        (B)
    \end{minipage}\\
    \begin{minipage}[t]{0.24\textwidth}
        \centering
        \includegraphics[width=1.0\linewidth]{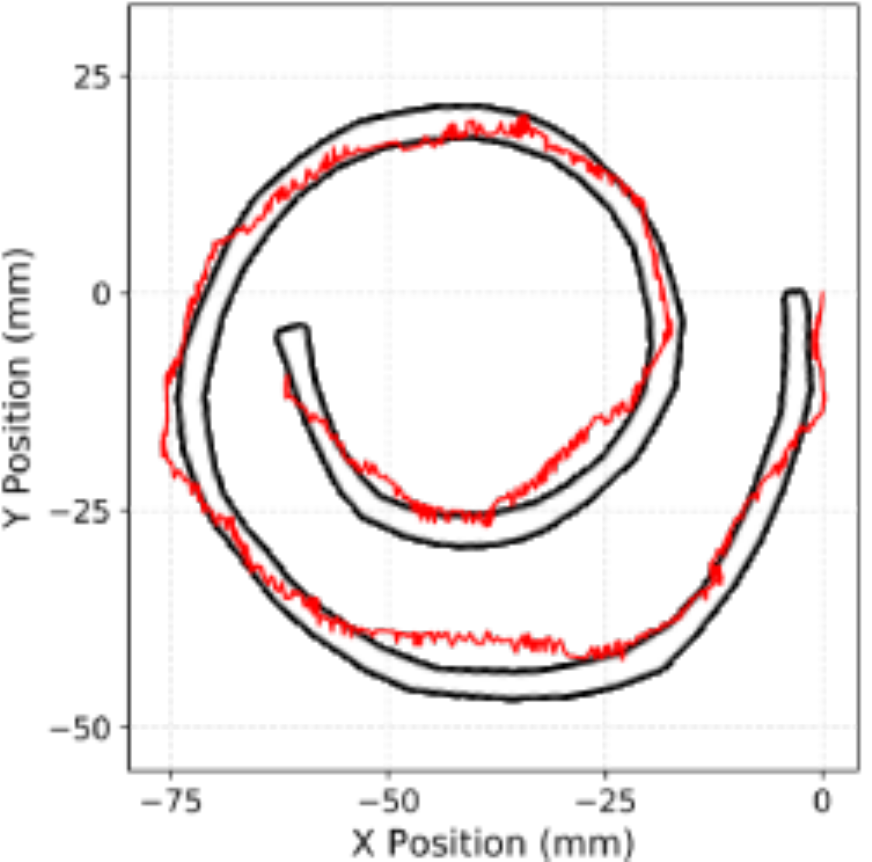}
        (C)
    \end{minipage}\hfill%
    \begin{minipage}[t]{0.24\textwidth}
        \centering
        \includegraphics[width=1.0\linewidth]{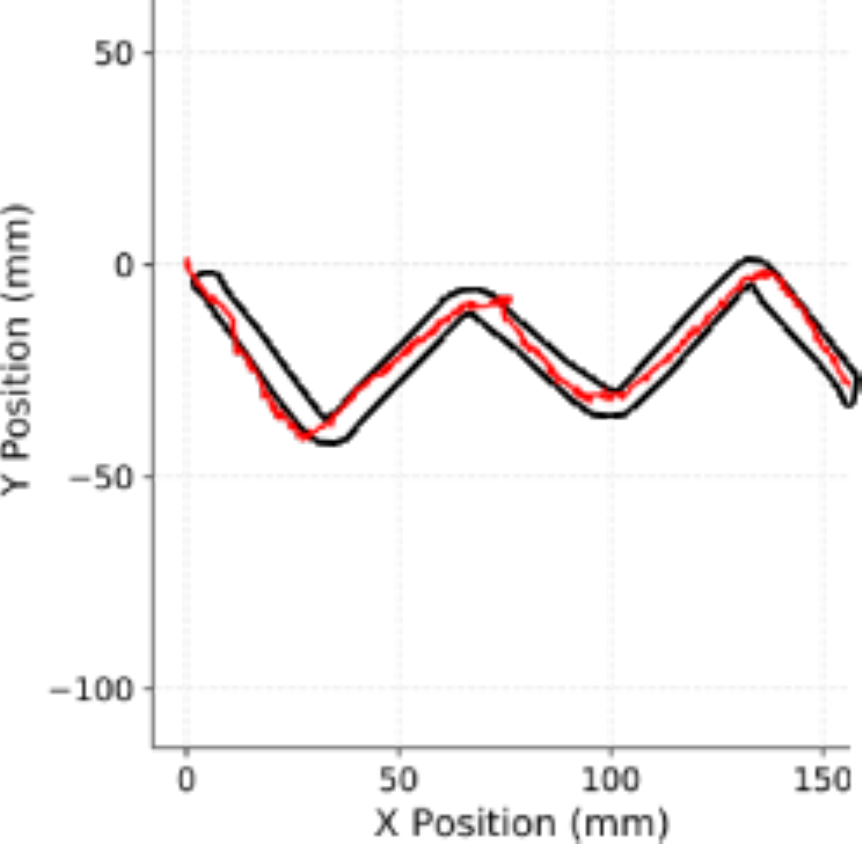}
        (D)
    \end{minipage}
    \caption{Edge Tracking Shapes, and Results. (A) Circular Edge, (B) Triangular Edge, (C) Spiral Edge, (D) Zig-Zag Edge}
    \label{fig:edge_tracking}
\end{figure}

As another implementation of our contour tracking pipeline, we demonstrate a simple controller that takes the contour location relative to the UR-10 manipulator, and outputs a motion vector for the finger to follow. The controller is based on a simple tactile servoing algorithm, where we try to maintain the location of the contour location in the center of the surface frame.

As the finger and the attached sensor move over the edge, only one portion of the \btsp is in contact with the edge surface. This can be detected and tracked by our controller, and since we start our controller execution with the sensor's center touching the edge, any deviations in the contours from this center is compensated by an opposing motion vector sent to the UR-10 manipulator as a control command.

We track the edges of various non-trivial patterns, namely circle, spiral, triangle, and zig-zag. We overlay the ground truth image of our shapes over the trajectory that is tracked from the robot's pose data for the finger. Barring minor alignment issues between the  surface and the finger, and some sliding experienced during the execution, the controller is able to guide the finger across the edges with relative accuracy.
The results, with the ground truth shapes, are shown in Figs.~\ref{fig:edge_tracking}. In each of the plots, we have the trajectory of the \btsp in world coordinate space in red, and the black polygons denote the inner and outer diameters of the edges of the shapes we track, also in world coordinate space measured in millimeters. We perform pixel-wise trajectory alignment to align the sensor pose to the ground truth boundary.
\section{Conclusion}
In conclusion, the work proposes a novel method to convert  raw tactile data from the \btsp sensor  into a spatio-temporal gradient (events) surface that closely tracks the regions of maximum tactile stimulus. Our algorithm approximates  the region of touch on the skin of the \btsp sensor sufficiently accurate  to perform various tactile feedback tasks. Specifically, we demonstrated the usefulness of the new representation experimentally for 
the tasks of tracking tactile stimulus across the sensor, measuring relative force, slippage detection and classification of direction, and tracking edges on a plane. 
%We demonstrate experiments and discuss results for each.
%We perform temporal aggregation on the generated events, and use that to obtain a Voronoi tesselation of the sensor's surface, which is then used to perform natural neighbors interpolation of the surface.
In comparison to other methods for tactile data processing, our method is real-time and requires minimal overhead in computation.
%, at the cost of highly accurate surface deformation values.

%The spatio-temporal surface, and the resulting contours, contain information about both the location and intensity of a tactile stimulus applied to the sensor's surface. Our intermediate contour representation can then be used to perform useful tasks, related to tracking tactile stimulus across the sensor, measuring relative force, slippage detection and classification of direction, and tracking edges on a plane. We demonstrate experiments and discuss results for each.

Lastly, our approach is independent of the particular sensor type, and we present an accompanying dataset of task-agnostic data samples gathered with the \btsp sensor. These include motion tracking over known trajectories and their time-synchronized RGB images, force sensor readings for varying forces applied to the surface using different indenter diameters, and slippage data for several objects and their accompanying ground truth timestamps.
%%%%%%%%%%%%%%%%%%%%%%%%%%%%%%%%%%%%%%%%%%%%%%%%%%%%%%%%%%%%%%%%%%%%%%%%%%%%%%%%
\bibliography{references}
\bibliographystyle{IEEEtran}

\end{document}